\documentclass[dvipsnames]{article} %
\usepackage{colm2024_conference}

\usepackage{booktabs}
\usepackage{enumitem}
\usepackage{wrapfig}
\usepackage{algorithm}
\usepackage{algpseudocode}
\usepackage{graphicx}
\usepackage[misc]{ifsym}

\usepackage{microtype}
\usepackage{amsmath}
\usepackage{colortbl}
\usepackage[utf8]{inputenc}
\usepackage[T1]{fontenc}
\definecolor{lightgray}{rgb}{0.9,0.9,0.9}
\usepackage{caption}
\usepackage{subcaption}
\usepackage{setspace}
\usepackage{url}
\usepackage{multirow}
\usepackage{tabularx}
\usepackage{blindtext}
\usepackage{pgfplots}
\pgfplotsset{compat=1.18} 
\usepackage{tikz}
\usetikzlibrary{er,positioning,bayesnet}
\usepackage{makecell}
\usepackage{tipa}
\usepackage{siunitx}
\usepackage{nicefrac}
\usepackage{listings}
\usepackage[raster,skins, most]{tcolorbox} %
\usepackage{xltabular}
\usepackage{adjustbox}
\usepackage{xurl}
\usepackage{rotating}
\usepackage[normalem]{ulem}
\usepackage{xspace}

\usepackage{fontawesome}

\usepackage{xcolor}      
\newcommand\inb[1]{\colorbox{gray!20}{\lstinline|#1|}}

\useunder{\uline}{\ul}{}

\usepackage{amsmath,amsfonts,bm}

\def\eqref#1{equation~\ref{#1}}

\def\1{\bm{1}}

\DeclareMathAlphabet{\mathsfit}{\encodingdefault}{\sfdefault}{m}{sl}
\SetMathAlphabet{\mathsfit}{bold}{\encodingdefault}{\sfdefault}{bx}{n}

\renewcommand{\ghlink}{https://github.com/Alibaba-NLP/DeepResearch}

\newcommand*\justify{%
  \fontdimen2\font=0.4em
  \fontdimen3\font=0.2em
  \fontdimen4\font=0.1em
  \fontdimen7\font=0.1em
  \hyphenchar\font=`\-
}

\renewcommand{\texttt}[1]{%
  \begingroup
  \ttfamily
  \begingroup\lccode`~=`/\lowercase{\endgroup\def~}{/\discretionary{}{}{}}%
  \begingroup\lccode`~=`[\lowercase{\endgroup\def~}{[\discretionary{}{}{}}%
  \begingroup\lccode`~=`.\lowercase{\endgroup\def~}{.\discretionary{}{}{}}%
  \catcode`/=\active\catcode`[=\active\catcode`.=\active
  \justify\scantokens{#1\noexpand}%
  \endgroup
}

\usepackage{makecell}
\usetikzlibrary{tikzmark}
\makeatletter
\newcommand*\myfontsize{%
  \@setfontsize\myfontsize{7}{8}%
}
\makeatother

\definecolor{uclablue}{RGB}{159, 195, 224}

\definecolor{uclagold}{RGB}{255, 240, 180}

\definecolor{aliceblue}{RGB}{255, 238, 241}

\definecolor{cadmiumgreen}{rgb}{0.0, 0.42, 0.24}

\definecolor{myred}{rgb}{0.7, 0.3, 0.0}
\definecolor{myblue}{rgb}{0.2, 0.3, 0.6}
\definecolor{babygreen}{rgb}{0.85, 0.97, 0.85}

\definecolor{purple1}{RGB}{126, 107, 196}
\definecolor{purple2}{RGB}{199, 158, 207}
\definecolor{purple3}{RGB}{214, 200, 255}
\definecolor{purple4}{RGB}{254, 240, 255}

\definecolor{deepblue}{RGB}{48, 58, 82}

\newcommand{\symboletongyi}{\raisebox{0pt}{~\includegraphics[scale=0.012]{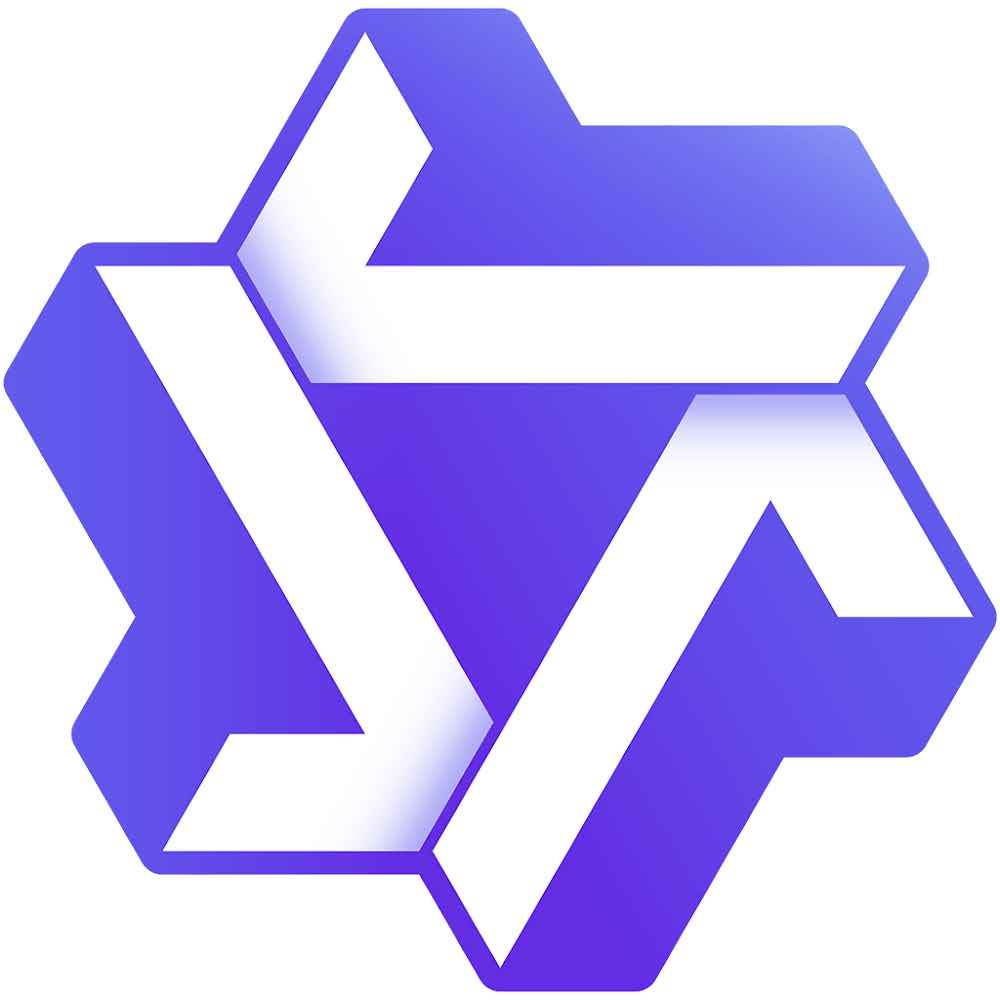}}~}

\definecolor{deepPurple}{HTML}{330066}

\definecolor{uclablue_old}{rgb}{0.15, 0.45, 0.68}
\hypersetup{
    breaklinks,
    citecolor=uclablue_old,
    colorlinks=true,
}

\newcommand{\papername}{WebResearcher\xspace}
\newcommand{\modelname}{IterResearch\xspace}
\newcommand{\dataname}{WebFrontier\xspace}

\newtcolorbox{mybox}[2][]
  {colback = black!5!white, colframe = black!75!black, fonttitle = \bfseries,
    colbacktitle = black!100!black, enhanced, before upper={\fontsize{8}{11}\obeyspaces\obeylines\selectfont}, fontupper=\selectfont,
    attach boxed title to top left={yshift=-2.2mm,xshift=4mm},
    title=#2,#1}

\title{%
\raisebox{-2.0em}{
  \parbox[t]{0.35in}{\includegraphics[width=0.6in]{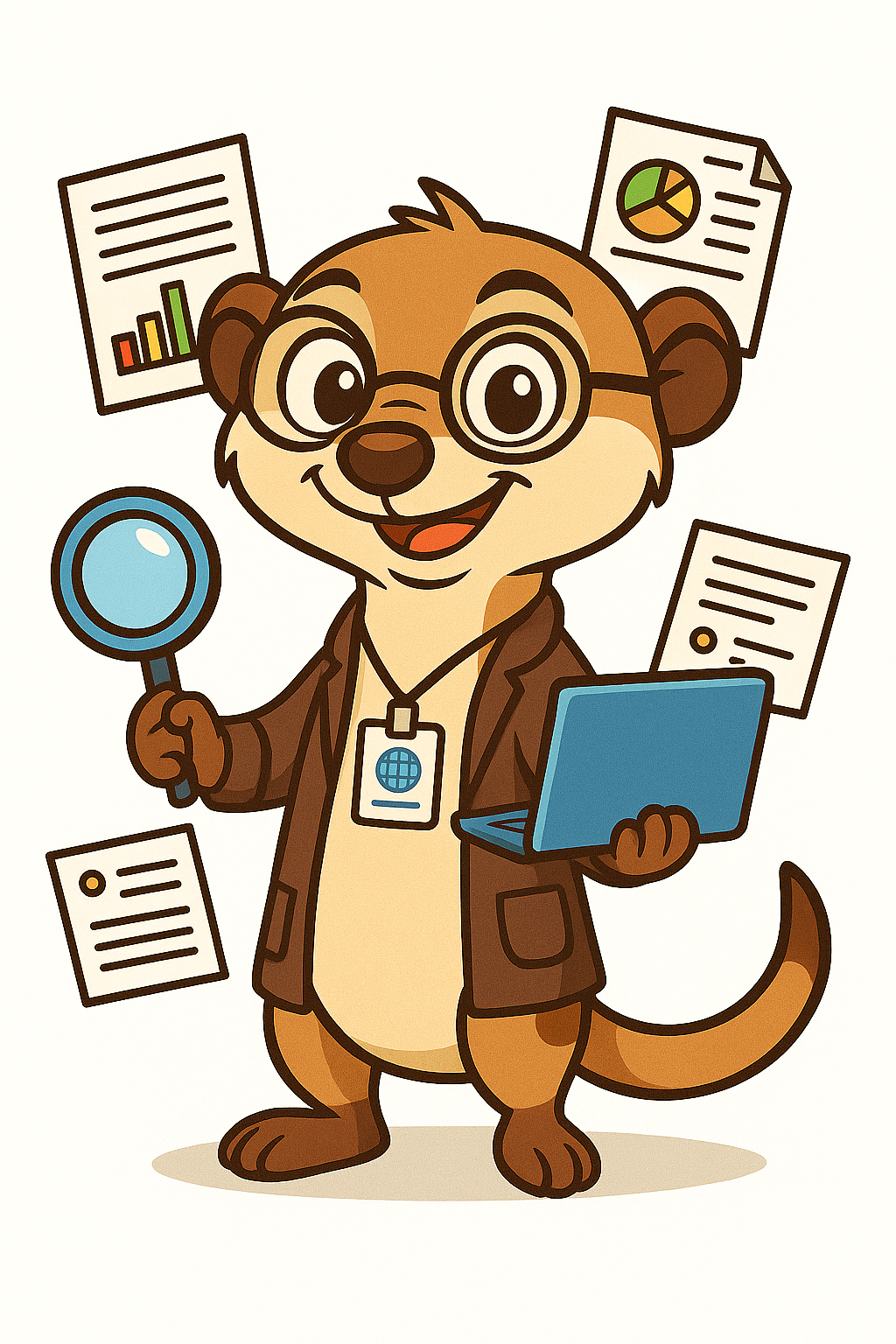}} 
  }
\begin{tabular}[t]{l} 
  \parbox[t]{0.8\textwidth}{\centering 
    WebResearcher: Unleashing unbounded reasoning capability in Long-Horizon Agents
  }
\end{tabular}
}

\author{%
\small{Zile Qiao$^{* (\textrm{\Letter})}$, Guoxin Chen$^{*}$, Xuanzhong Chen$^{*}$, Donglei Yu\thanks{Equal Core Contributors.}\hspace{0.5mm}, Wenbiao Yin, Xinyu Wang, Zhen Zhang, Baixuan Li, Huifeng Yin, Kuan Li, Rui Min, Minpeng Liao, Yong Jiang$^{(\textrm{\Letter})}$, Pengjun Xie, Fei Huang, Jingren Zhou}%
  \\[1em]               
  {\fontsize{10pt}{11pt}\selectfont          
Tongyi Lab\symboletongyi, Alibaba Group}\\
}

\begin{document}

\maketitle

\begingroup
  \renewcommand\thefootnote{\Letter}  
  \footnotetext{Corresponding authors. \{qiaozile.qzl, yongjiang.jy\}@alibaba-inc.com} 
\endgroup

\vspace{-1em}
\begin{abstract}
Recent advances in deep-research systems have demonstrated the potential for AI agents to autonomously discover and synthesize knowledge from external sources.
In this paper, we introduce \textbf{\papername}, a novel framework for building such agents through two key components:
(1) \textbf{\modelname}, an iterative deep-research paradigm that reformulates deep research as a Markov Decision Process, where agents periodically consolidate findings into evolving reports while maintaining focused workspaces—overcoming the context suffocation and noise contamination that plague existing mono-contextual approaches; and (2) \textbf{\dataname}, a scalable data synthesis engine that generates high-quality training data through tool-augmented complexity escalation, enabling systematic creation of research tasks that bridge the gap between passive knowledge recall and active knowledge construction.
Notably, we find that the training data from our paradigm significantly enhances tool-use capabilities even for traditional mono-contextual methods.
Furthermore, our paradigm naturally scales through parallel thinking, enabling concurrent multi-agent exploration for more comprehensive conclusions.
Extensive experiments across 6 challenging benchmarks demonstrate that \papername achieves state-of-the-art performance, even surpassing frontier proprietary systems.
\end{abstract}

\begin{figure*}[h]
\centering
\begin{subfigure}[b]{0.45\textwidth}
    \centering
    \includegraphics[width=\linewidth]{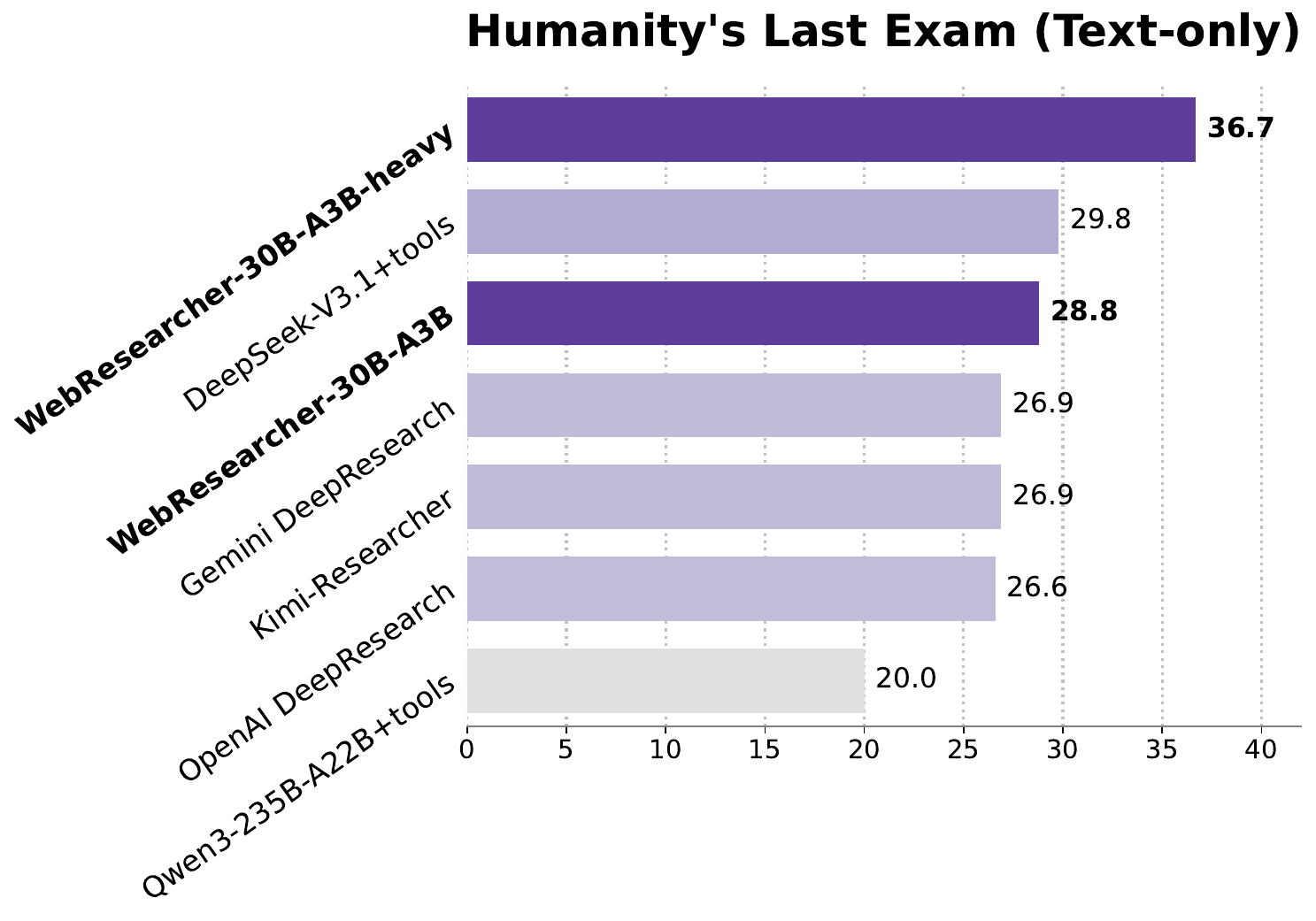}
    \label{fig:hle_res}
\end{subfigure}
\begin{subfigure}[b]{0.45\textwidth}
    \centering
    \includegraphics[width=\linewidth]{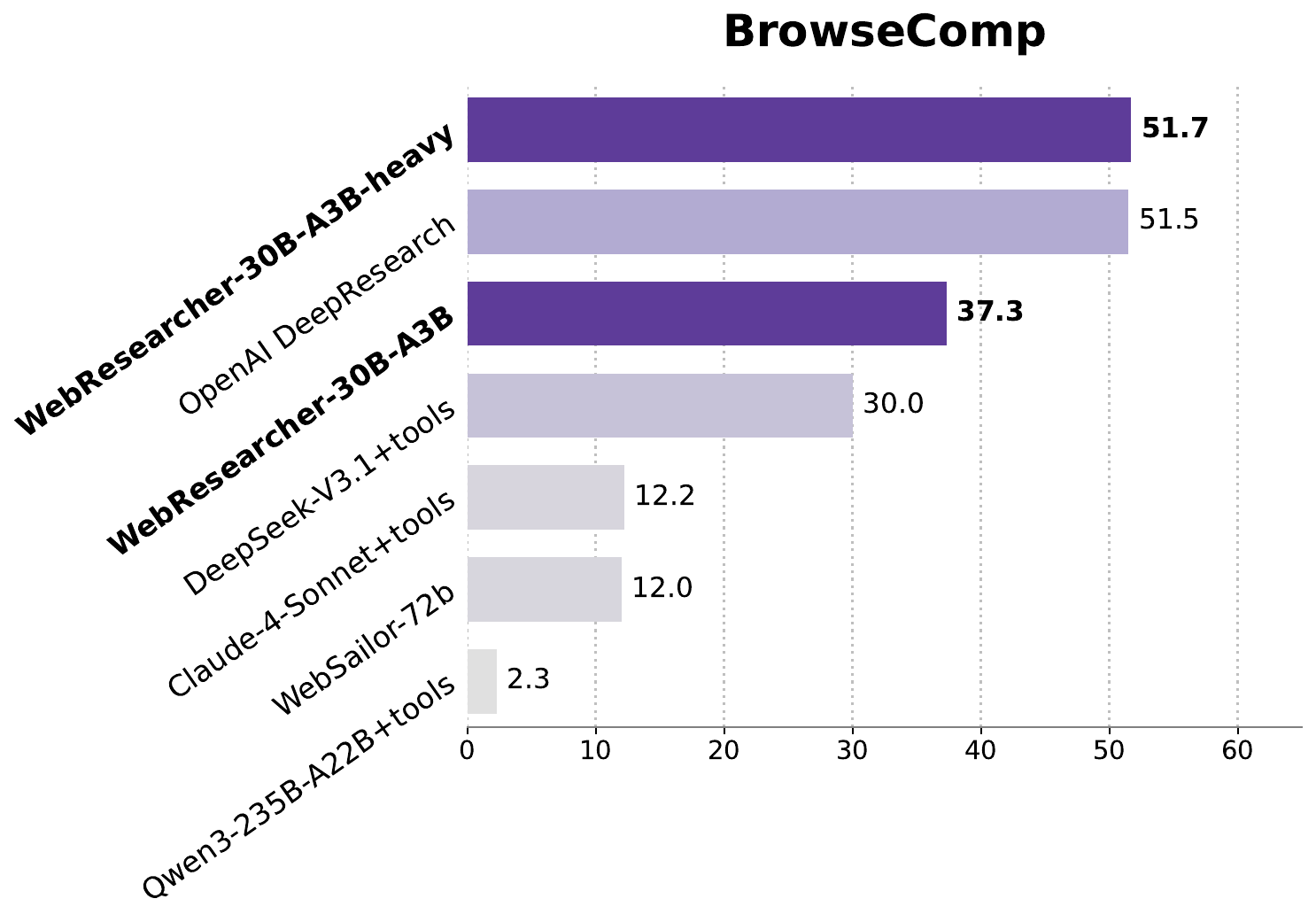}
    \label{fig:hle_res}
\end{subfigure}
\caption{Performance comparison between \papername and state-of-the-art deep-research agents.}
\label{fig:main_results}
\end{figure*}

\newpage

\newpage
\section{Introduction}
\label{sec:intro}

The pursuit of artificial general intelligence (AGI) has historically focused on scaling models to acquire vast passive knowledge~\citep{claude,gemma3,gemini_25_pro,llama4,openai_o3_o4_mini,guo2025deepseek,yang2025qwen3}.
However, this knowledge-centric approach may reach a critical limitation: while models can memorize and recall information, they struggle to actively discover, verify, and synthesize new knowledge from external sources—a capability fundamental to human intelligence.
This limitation has catalyzed a paradigm shift toward actively autonomous agent systems that emulate human research workflows. Rather than relying solely on pre-trained knowledge, these systems dynamically construct understanding by autonomously decomposing complex problems, orchestrating sophisticated tool use, and synthesizing disparate findings into coherent, evidence-grounded narratives. This emerging class of systems, often referred to as deep research~\citep{dr}, represents a crucial step toward AGI by bridging the gap between passive knowledge repositories and active knowledge constructors.

Recent advances in deep-research systems, exemplified by pioneers like OpenAI's Deep Research~\citep{dr}, Google's Gemini Deep Research~\citep{google_dr}, Grok DeepSearch~\citep{grok3}, and Kimi-Researcher~\citep{kimi-researcher}, have demonstrated breakthrough performance on challenging benchmarks including Humanity's Last Exam (HLE)~\citep{hle} and BrowseComp~\citep{bc_en}.
The success of these proprietary systems has spurred significant open-source development.
Recent open-source efforts including WebThinker~\citep{Li2025webthinker}, WebShaper~\citep{tao2025webshaper}, and WebSailor~\citep{li2025websailor} have shown competitive performance in deep-research tasks.
Notably, these open-source implementations have converged on a remarkably similar architectural pattern: a \textit{mono-contextual paradigm} that continuously accumulates all retrieved information and intermediate reasoning steps into a single, ever-expanding context window~\citep{chen2025cpo,jin2025search,li2025search,wu2025webdancer}. While this linear accumulation strategy appears intuitive and has shown initial success, a deeper analysis reveals that it fundamentally constrains the potential of deep-research agents.
Specifically, this prevalent paradigm suffers from two critical limitations that become increasingly severe as research complexity grows:
\textbf{(1) Cognitive Workspace Suffocation}: The ever-expanding context progressively constrains the model's capacity for deep reasoning, as the fixed context window becomes dominated by accumulated data rather than active thinking space, forcing premature conclusions.
\textbf{(2) Irreversible Noise Contamination}: Without mechanisms to filter or revise earlier content, irrelevant information and initial errors persist throughout the entire process, diluting signal quality and propagating biases that compound over time.
These limitations reveal a paradox: as deep-research agents gather more information to solve complex problems, their mono-contextual architecture becomes increasingly ineffective at processing and reasoning over that very information.

In this work, we introduce \textbf{\modelname}, an \textit{Iterative Deep-Research Paradigm} that reformulates deep research as a Markov Decision Process (MDP)~\citep{bellman1957markovian,puterman1990markov}.
Unlike the mono-contextual approach that suffers from unbounded state expansion and noise contamination, \modelname periodically consolidates its findings into a synthesized report and reconstructs its workspace, maintaining both continuity of knowledge and clarity of reasoning at arbitrary depths of research.
Specifically, \modelname operates through discrete rounds where each state contains only essential components: the research question, an evolving report synthesizing all previous findings and current research progress, and the immediate context from the recent tool interaction.
This evolving report serves as the agent's central memory—progressively refined through each round as new insights are integrated with existing knowledge.
Between rounds, a state transition function preserves this updated report while discarding ephemeral information, ensuring the Markov property while preventing information loss.
This periodic synthesis is the core of our paradigm: it not only preserves essential knowledge to guide subsequent reasoning but also maintains a focused cognitive workspace for each phase, effectively preventing both suffocation and noise propagation.
Therefore, \modelname achieves what mono-contextual systems cannot—sustained high-quality reasoning across the entire research process, enabling the agent to pursue arbitrarily complex investigations through iterative refinement rather than exhaustive single-pass accumulation.

Furthermore, to address the critical bottleneck of data scarcity in training deep-research agents, we develop \textbf{\dataname}, a \textit{Scalable Data Synthesis Engine} that leverages large language models augmented with diverse external tools to systematically generate high-quality training data for complex research tasks.
\dataname tackles a fundamental challenge in agentic AI development: how to create high-quality and large-scale training data while maintaining factual accuracy and verifiability.
Our approach employs a three-stage iterative workflow—seed generation from diverse corpora, tool-augmented complexity escalation, and rigorous quality control—to produce tasks that effectively span the capability gap between baseline models and their tool-augmented counterparts.
The engine's core mechanism involves a self-bootstrapping process where tool-augmented agents progressively refine simple questions into research problems that require multi-source synthesis, cross-domain reasoning, and computational verification.
This systematic approach enables the generation of a large-scale dataset that explore varying levels of complexity while ensuring factual grounding.
The synthesized data serves as the foundation for training \modelname through a multi-stage training, enabling the model to acquire both robust tool-use capabilities and sophisticated reasoning skills.

Finally, at inference time, we introduce the \textbf{Research-Synthesis Framework}, built upon our IterResearch paradigm. This framework consists of two phases: Parallel Research and Integrative Synthesis. In the Parallel Research phase, multiple Research Agents concurrently solve the target problem following the IterResearch method, with each agent deriving a final report and the predicted answer. Subsequently, in the Integrative Synthesis phase, a single Synthesis Agent integrates these findings to produce a more comprehensive and robust conclusion. By synthesizing from the final reports rather than the entire research trajectories, the Synthesis Agent can process a wider diversity of research paths within a constrained context. This approach effectively leverages test-time scaling, maximizing the benefits of divergent exploration in complex deep research scenarios.

We systematically evaluate \papername on benchmarks spanning diverse domains and task types. Our experiments demonstrate that \papername achieves state-of-the-art performance across 6 challenging benchmarks, even surpassing frontier proprietary systems. On Humanity's Last Exam (HLE), one of the most demanding tests for AI reasoning, \papername-heavy achieves 36.7\% accuracy, substantially outperforming all existing systems including DeepSeek-V3.1 (29.8\%) and OpenAI Deep Research (26.6\%). Similarly, on complex web navigation tasks like BrowseComp-en, our system reaches 51.7\%, matching OpenAI's proprietary Deep Research system while exceeding the best open-source alternative by 21.7 percentage points. These results validate that our iterative synthesis paradigm fundamentally addresses the limitations of mono-contextual approaches, enabling sustained high-quality reasoning even in the most complex research scenarios. 
Furthermore, we demonstrate that the training data generated under the \modelname paradigm \textbf{provides benefits beyond our own system}. When traditional mono-contextual methods are trained with our iterative paradigm data, they show substantial performance improvements. 
This highlights that our paradigm produces superior training signals that enhance agentic capabilities.

\newpage
\section{\modelname: An Iterative Deep-Research Paradigm}
\begin{figure}[h]
    \centering
    \includegraphics[width=\linewidth]{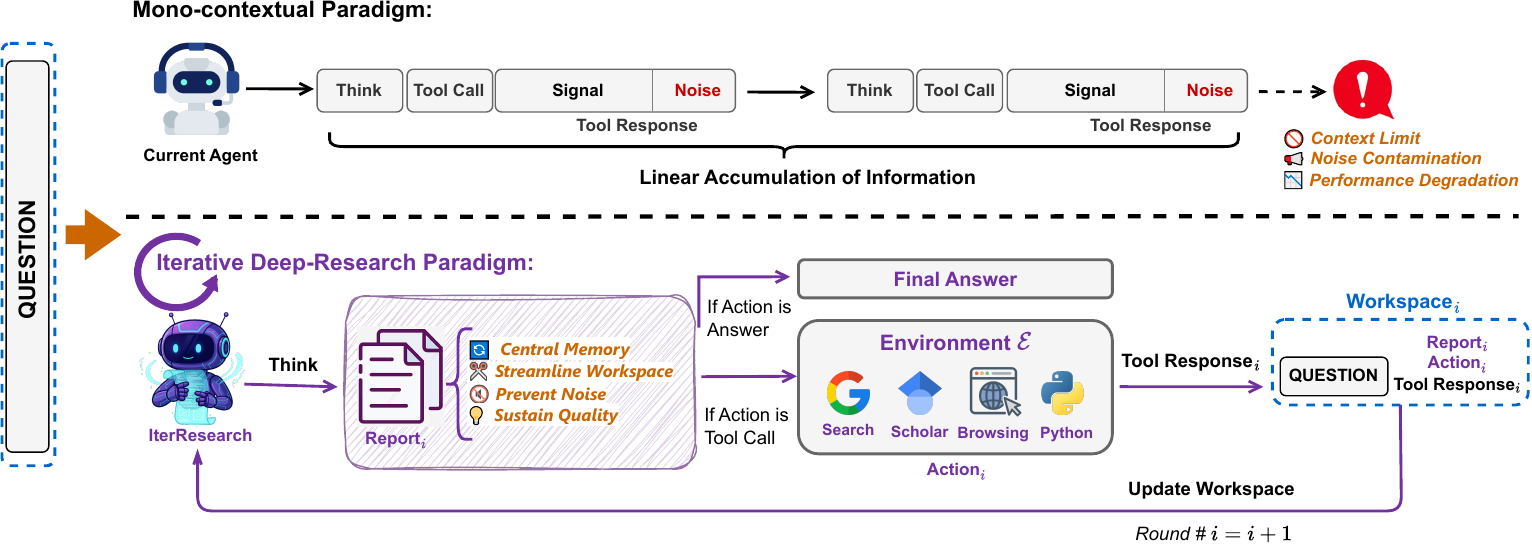}
    \vspace{-2em}
    \caption{
    An illustration of our \textbf{Iterative Deep-Research Paradigm} in contrast to the prevalent \textbf{Mono-contextual Paradigm}.
    \textbf{(Top)} The mono-contextual approach linearly accumulates all information into a single, ever-expanding context, leading to cognitive suffocation and noise contamination.
    \textbf{(Bottom)} Our \modelname paradigm deconstructs research into discrete rounds. In each round $i$, the agent operates on a lean, reconstructed \texttt{Workspace}. It first \textbf{Thinks}, then \textbf{synthesizes} new findings into an evolving summary \textbf{Report$_i$}, and finally decides on an \textbf{Action$_i$}.
    The crucial step is the reconstruction: the \texttt{Workspace} for the next round is rebuilt using only the essential outputs of the previous round (the updated \texttt{Report} and \texttt{Tool Response}), thus preventing context bloat and enabling sustained reasoning.}
    \label{fig:IterResearch}
\end{figure}

Deep-research agents aim to accomplish in minutes what would take human researchers hours. Such long-horizon tasks require navigating heterogeneous evidence sources, coordinating multiple rounds of tool use, and maintaining coherent chains of reasoning across an ever-expanding body of information.
Yet, this very complexity directly challenges the mono-contextual, linear accumulation paradigm adopted by current research agents~\citep{Li2025webthinker,tao2025webshaper,li2025websailor}, which is fundamentally constrained by:
(1) Cognitive Workspace Suffocation: As the context window fills with accumulated data, the model's capacity for active reasoning diminishes. The fixed context budget becomes dominated by historical information rather than providing space for deep thinking, forcing premature conclusions when the window approaches its limit.
(2) Irreversible Noise Contamination: Without mechanisms to filter or revise earlier content, irrelevant information and initial errors persist throughout the entire process. This noise accumulation dilutes signal quality and propagates biases that compound over time, degrading the overall research quality.
These limitations create a paradox: as agents gather more information to solve complex problems, their mono-contextual architecture becomes increasingly ineffective.

To overcome these limitations, we propose \modelname, which reformulates deep research as a Markov Decision Process (MDP) with periodic state reconstruction.
Instead of maintaining an ever-expanding context, \modelname operates through discrete rounds where each state contains only essential components, as illustrated in Figure~\ref{fig:IterResearch}.
The key insight of \modelname is to replace linear accumulation with iterative synthesis and reconstruction. Each research round operates on a focused \texttt{Workspace} that maintains clarity while preserving continuity through an evolving report that serves as the agent's central memory.
At each round $i$, the agent's state $s_i$ consists of three components: (1) The original research \texttt{Question} $q$, (2) The evolving \texttt{Report}$_{i-1}$ from the previous round (empty for $i=1$), and (3) The most recent \texttt{Action}$_{i-1}$ and its \texttt{Tool Response}$_{i-1}$ (if $i>1$).
This compact state representation ensures the Markov property while maintaining all essential information for decision-making.

To implement this iterative paradigm effectively, we define three structured meta-information categories—\inb{Think}, \inb{Report}, and \inb{Action}—that guide the agent's decision in each round:
\begin{itemize}
    \item \inb{Think}: This component serves as the agent's cognitive scratchpad where it articulates its internal reasoning process. The agent analyzes the current state (workspace), evaluates the outcome of its previous action, reflects on the research progress, and formulates a plan for its next action. This component ensures the agent's decision-making is transparent and interpretable for current state, and is not directly used in subsequent rounds to prevent clutter.

    \item \inb{Report}: The centerpiece of our paradigm, this component represents the agent's evolving central memory.
    Rather than appending raw data, the agent synthesizes new findings with existing knowledge to produce a coherent, high-density summary.
    This updated report captures all critical information discovered to date and serves as the primary component for constructing the next round's workspace.
    
    \item \inb{Action}: The agent's concrete action for the current round, which takes one of two forms:
    \begin{itemize}
        \item \textit{Tool Call}: A specific command to interact with the external \texttt{Environment}, such as invoking a search engine or a code interpreter, to gather new information.
        \item \textit{Final Answer}: A terminal action, generated when the agent determines it has sufficient evidence to resolve the initial \texttt{Question}. This concludes the research process.
    \end{itemize}
\end{itemize}

Our \modelname paradigm fundamentally reimagines deep research as an iterative synthesis process rather than linear accumulation. The complete research unfolds through discrete rounds: starting from just the research question, the agent generates its initial Think-Report-Action triplet; in subsequent rounds, it reconstructs a focused workspace from the question, previous report, and latest tool response, then produces an updated synthesis. This Report synthesis is the cornerstone of our approach—the agent doesn't merely append new findings but actively integrates them with existing knowledge, resolving conflicts and updating conclusions to maintain a coherent, high-density summary that captures all critical discoveries while filtering out noise. The process continues until the agent determines  sufficient evidence has been gathered, producing a Final Answer.

This iterative paradigm provides structural advantages that compound over long-horizon research. By maintaining a constant-size workspace regardless of research depth, \modelname preserves full reasoning capacity throughout the entire process—where mono-contextual systems suffer from diminishing returns as contexts bloat, our approach maintains consistent performance whether conducting ten or hundred rounds of investigation. The periodic synthesis acts as an intelligent filter, preserving signal while eliminating noise, enabling error recovery through report revision, and ensuring monotonic information gain. Through this disciplined state maintenance centered on evolving reports, \modelname transforms deep research from exhaustive single-pass accumulation into iterative refinement, achieving theoretically unbounded research depth while maintaining both efficiency and quality—capabilities that are fundamentally impossible under the mono-contextual paradigm.

\section{A Scalable Data Engine for Advancing Agentic Intelligence}
\label{sec:data}

\begin{figure}[h]
\centering
\includegraphics[width=\linewidth]{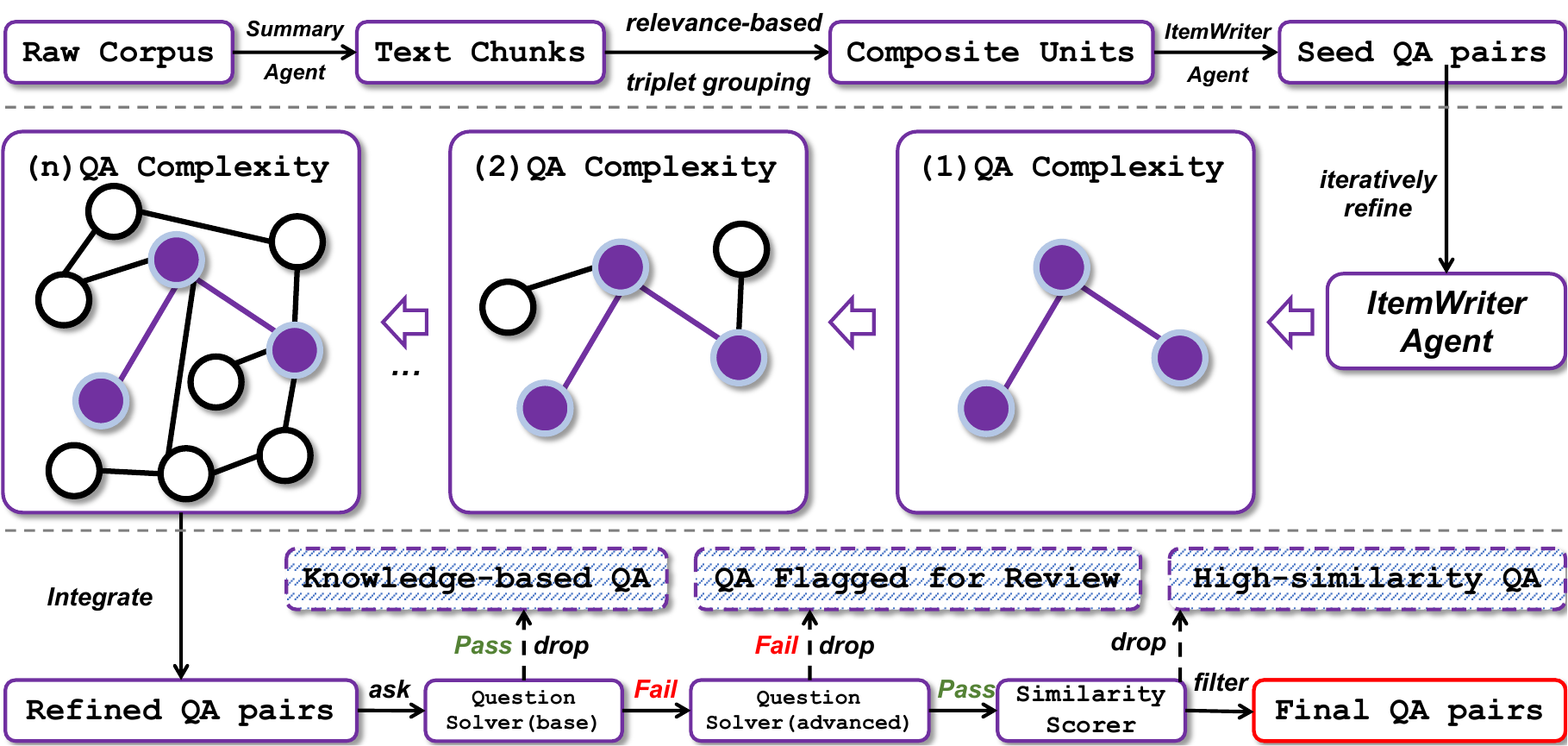}
\caption{Overview of the three-stage data synthesis workflow, powered by a multi-agent system. The process begins with seed data generation from a curated corpus. It then enters an iterative loop where tool-augmented agents systematically increase task complexity. The workflow concludes with a multi-stage quality control process to calibrate difficulty and ensure factual correctness.}
\label{fig:data_engine}
\end{figure}

The advancement of agentic intelligence, characterized by capabilities of complex reasoning and autonomous tool use, is fundamentally constrained by the quality and complexity of their training data. To address this limitation, we introduce a scalable data engine designed to synthesize a large-scale, high-quality dataset that systematically probes and extends the capabilities of current models. Our engine leverages a collaborative multi-agent framework organized into a three-stage iterative workflow: (1) seed data generation, (2) iterative complexity escalation, and (3) rigorous quality control. As depicted in Figure~\ref{fig:data_engine}, this process orchestrates a team of specialized agents to generate progressively more challenging tasks.

\subsection{Stage 1: Seed Data Generation}
The process initiates with a diverse, multidisciplinary corpus of contemporary documents, including webpages, academic papers, and e-books. A Summary Agent preprocesses this corpus by paraphrasing content, removing artifacts (e.g., HTML tags), and distilling the text into information-dense chunks.

To generate initial tasks that require non-trivial reasoning, we form composite units by combinatorially grouping these thematically related chunks. An ItemWriter Agent is then prompted with these composite units to generate seed question-answer (QA) pairs. These initial pairs are designed to require multi-source information synthesis, thereby providing the foundation for the subsequent complexity escalation stage.

\subsection{Stage 2: Iterative Complexity Escalation}
The core of the data engine is a self-bootstrapping refinement loop orchestrated by the ItemWriter Agent. At this stage, the agent is augmented with a suite of external tools: (i) general web search, (ii) academic literature search, (iii) webpage browswer, and (iv) Python code interpreter. For each seed QA pair, the tool-augmented agent iteratively evolves both the question and answer to increase their cognitive complexity and extend their scope beyond the original context. This iterative evolution is driven by four key operations. Initially, the agent performs knowledge expansion, querying external sources to broaden the question's scope. It then engages in conceptual abstraction, analyzing materials to distill higher-level principles and identify subtle cross-domain relationships. To ensure correctness, factual grounding is achieved through multi-source cross-validation, enhancing the answer's accuracy and depth. Finally, the agent leverages a Python environment for computational formulation, crafting problems that demand quantitative calculation or logical simulation.

This iterative process creates a virtuous cycle where a more sophisticated QA pair generated in one iteration becomes the seed for the next. This enables a controlled and systematic escalation in task complexity.

\subsection{Stage 3: Rigorous Quality Control}
To ensure the final dataset is of high quality and precisely calibrated to the target difficulty, all generated QA pairs undergo a rigorous validation process managed by specialized agents. First, a QuestionSolver Agent, operating in a baseline mode without access to tools, attempts to answer each question. Any pair answered correctly at this step is deemed too simple for our target complexity level and is filtered out. Second, the remaining challenging pairs are passed to the same QuestionSolver Agent, now operating in an advanced, tool-augmented mode that mirrors the capabilities of our target model. Pairs that the agent successfully solves in this mode are designated as high-value, complex-reasoning instances and retained for the final dataset. Conversely, any pair that this advanced agent fails to solve is considered intractable or potentially flawed, and is consequently discarded or flagged for expert human review. Throughout this validation pipeline, a Judge Agent automatically assesses the correctness of the solver's output against the ground-truth answer. Concurrently, a SimilarityScorer Agent filters out newly generated pairs that are semantically redundant with existing data, thereby maintaining dataset diversity.

In summary, our data engine is designed to achieve three primary objectives: (1) efficiently generate a large volume of complex tasks situated within the "capability gap" between a baseline model and its tool-augmented counterpart; (2) ensure that all generated tasks maintain high complexity while being factually correct and verifiable; and (3) systematically map and expand the frontiers of reasoning and tool-use for advanced LLM agents.

\section{Training and Test-Time Optimization}
\label{sec:training}

\subsection{Rejection Sampling Fine-Tuning}

To train \modelname, we adopt a rejection sampling fine-tuning (RFT) approach that leverages well-formed trajectories generated by prompting large language models to follow our iterative paradigm's structured format.

\textbf{Trajectory Generation and Filtering.} For each training instance consisting of a research question $q^{(i)}$ and reference answer $a^{(i)}$, we employ prompt the LLMs to generate multiple research trajectories following the \modelname paradigm. Each trajectory $\tau^{(i)} = \{(s_1^{(i)}, r_1^{(i)}, o_1^{(i)}), ..., (s_{T_i}^{(i)}, r_{T_i}^{(i)}, o_{T_i}^{(i)})\}$ consists of $T_i$ rounds, where $s_j^{(i)}$ represents the state at round $j$, $r_j^{(i)}$ represents the structured response (Think-Report-Action) and $o_j^{(i)}$ denotes the corresponding tool observation. We apply strict rejection sampling, retaining only trajectories whose final answers exactly match the reference $a^{(i)}$, ensuring the training data embodies both correct reasoning processes and accurate conclusions.

\textbf{Training Objective.} The model learns to generate structured responses conditioned on the iterative research context. Specifically, at each round $j$, the model must produce $r_j^{(i)}$ given the current state $s_j^{(i)}$. The training objective maximizes the conditional log-likelihood across all accepted trajectories:
\begin{equation}
\mathcal{L}(\theta) = \sum_{i=1}^{K} \sum_{j=1}^{T_i} \log p_\theta \left( r_j^{(i)} \,\Big|\, s_{j-1}^{(i)} \right),
\end{equation}

where $K$ denotes the number of accepted trajectories and $\theta$ represents model parameters. Crucially, this objective enforces the Markov property of our paradigm—each round's generation depends only on the immediate previous state rather than the entire history. During training, we compute gradients only over the model-generated response tokens $r_j^{(i)}$, treating observations $o_j^{(i)}$ as given context. This ensures the model learns to reason and synthesize rather than to predict tool outputs, maintaining a clear separation between the reasoning agent and external tools.

\subsection{Reinforcement Learning}
To further enhance \modelname's research capabilities, we employ reinforcement learning to optimize the model's ability to explore diverse reasoning paths while maintaining high-quality synthesis at each round.
A key advantage of our iterative paradigm is that each trajectory naturally decomposes into multiple training samples—one for each research round—whereas mono-contextual approaches yield only a single sample per trajectory. Specifically, for each research question $q^{(i)}$ with $G$ rollouts, trajectory $g$ unfolds over $T_g^{(i)}$ rounds, where each round $j$ produces a training tuple $(s_{g,j}^{(i)}, r_{g,j}^{(i)}, o_{g,j}^{(i)})$ consisting of the state, response, and tool response.
This decomposition yields a rich training corpus:
\begin{equation}
\mathcal{C}^{(i)} = \left\{ (s_{g,j}^{(i)}, r_{g,j}^{(i)}) : g \in [1,G], j \in [1,T_g^{(i)}] \right\},
\end{equation}
containing $\sum_{g=1}^G T_g^{(i)}$ samples per question. Aggregating across all $N$ training questions, our iterative paradigm generates a total corpus:
\begin{equation}
\mathcal{C}_{\text{total}} = \bigcup_{i=1}^{N} \mathcal{C}^{(i)}, \quad |\mathcal{C}_{\text{total}}| = \sum_{i=1}^{N} \sum_{g=1}^G T_g^{(i)}.
\end{equation}
This represents a substantial data amplification compared to mono-contextual approaches which would yield only $N \times G$ samples.
However, the variable trajectory lengths introduce a practical challenge: the total sample count varies across batches due to different $T_g^{(i)}$ values, conflicting with distributed training requirements for fixed batch sizes.
To address this while preserving data efficiency, we employ \textit{minimal-loss downsampling}, reducing the entire training corpus to the largest multiple of the data parallel (DP) size that does not exceed the original count:
\begin{equation}
|\mathcal{C}_{\text{train}}| = \left\lfloor \frac{|\mathcal{C}_{\text{total}}|}{\text{DP}_{\text{size}}} \right\rfloor \times \text{DP}_{\text{size}}.
\end{equation}
This approach ensures uniform distribution across devices while minimizing data loss (typically <1\%), maintaining distributed training stability.

To optimize \modelname over these multi-round trajectories, we adopt Group Sequence Policy Optimization (GSPO)~\citep{zheng2025group}.
We optimize the following objective:
\begin{equation}
    \mathcal{J}_{\text{GSPO}}(\theta) = \mathbb{E}_{q \sim \mathcal{Q}, \mathcal{C}_{\text{train}}\sim \pi_{\theta_{\text{old}}}(\cdot|q)} \left[\frac{1}{|\mathcal{C}_{\text{train}}|}\sum_{g=1}^G\sum_{j=1}^{T_g} \min(\rho_{g,j}(\theta)\hat{A}_{g,j}, \text{clip}(\rho_{g,j}(\theta), 1-\varepsilon, 1+\varepsilon)\hat{A}_{g,j})\right]
\end{equation}
where $\mathcal{Q}$ is the training set, $\hat{A}_{g,j} = \frac{r_{g,j} - \mu_r}{\sigma_r}$ is the normalized advantage, with $\mu_r$ and $\sigma_r$ computed across all $(g,j)$ pairs in $\mathcal{C}_{\text{train}}$, and $\rho_{g,j}(\theta)$ is the importance ratio based on sequence likelihood.
Notably, all $\sum_{g=1}^G T_g$ rounds from the $G$ trajectories for question $q$ form \textit{one group}, enabling efficient batched training while respecting the variable-length nature of our iterative research process. This differs from traditional GSPO where each trajectory would be treated separately—our approach leverages the natural decomposition of trajectories into rounds, treating each round as an independent training sample while maintaining group-level advantage normalization across all rounds. This design maximizes data utilization and ensures balanced learning across different research depths.

\subsection{Research-Synthesis: Harnessing Test-time Scaling with IterResearch}
\label{sec:tts}

To further unlock the potential of IterResearch, we further investigate test-time scaling. Given that DeepResearch involves multi-round tool calls and intensive reasoning, directly aggregating the context from every complete trajectory is computationally infeasible. Therefore, effective context management during test-time scaling is crucial, enabling the use of minimal context to accurately represent the problem-solving logic of a trajectory.

\begin{figure}
    \centering
    \includegraphics[width=\linewidth]{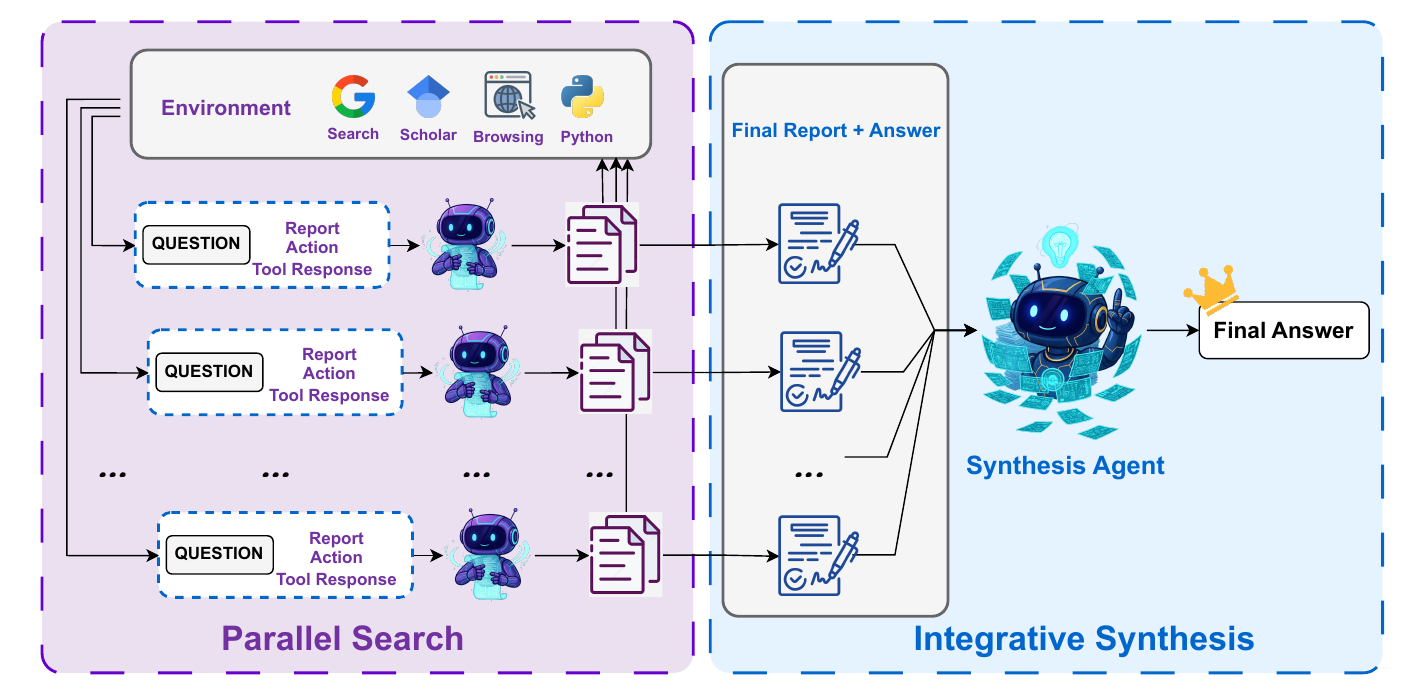}
    \caption{Illustration of Reason-Synthesis Framework}
    \label{fig:last-k-fusion}
\end{figure}

To address this challenge, we introduce the \textbf{Research-Synthesis Framework} as illustrated in Figure \ref{fig:last-k-fusion}. The framework consists of two distinct phases: Parallel Research and Integrative Synthesis. The former phase fosters the concurrent exploration of diverse problem-solving approaches, while the latter integrates these disparate perspectives into a single, unified solution.

\paragraph{Parallel Research} In Parallel Research phase, we employ $n$ Research Agents to independently solve the target problem. Each agent adheres to the IterResearch paradigm but carves out a unique solution trajectory by invoking distinct tools and generating different lines of reasoning. Ultimately, this phase yields a set of final reports and their corresponding predicted answers, one from each agent. This collection can be formally expressed as:
\begin{equation}
\begin{split}
    & \mathcal{M} = \{(\mathtt{Final\_Report}_u, \mathtt{Answer}_u) : u \in [1,n]\} \\
    & (\mathtt{Final\_Report}_u, \mathtt{Answer}_u) = \mathrm{IterResearch}_u(q) 
\end{split}    
\end{equation}

\paragraph{Integrative Synthesis} The Integrative Synthesis phase employs a single Synthesis Agent to consolidate the findings from all Research Agents and produce a final, reasoned conclusion. It takes the complete set of reports and answers as input to generate the final answer, represented as: 
\begin{equation}
    \mathtt{Final\_Answer} = \mathrm{Synthesis}(\mathcal{M})
\end{equation}

It is noted that each report from IterResearch concisely encapsulates its entire reasoning path. Consequently, the Synthesis Agent can assess a broader range of solution strategies under a limited context, fully harnessing the power of test-time scaling. In our experiments, we employ Qwen3-235B-A22B as our Synthesis Agent.

\newpage
\section{Experiments}

\subsection{Experimental Setup}
\paragraph{Models and Benchmarks} We implement our \papername using Qwen3-30B-A3B~\citep{yang2025qwen3} as the backbone model, considering both model performance and computational efficiency.
The complete system integrates our iterative research paradigm (\modelname) with web-scale training data constructed via \dataname.
To comprehensively evaluate \papername's capabilities, we conduct extensive experiments on 8 challenging benchmarks:
\begin{itemize}
    \item \textbf{HLE}~\citep{hle} - Humanity's Last Exam is an expert-curated benchmark of 2,500 highly challenging questions spanning a wide range of disciplines, designed to assess frontier-level academic competence. We use the 2,154 text-only questions.
    \item \textbf{GAIA}~\citep{mialon2023gaia} - A set of 466 real-world task questions evaluating general AI assistants under demanding conditions, emphasizing multi-step reasoning, multimodality, and tool use. We adopt 103 cases from the text-only validation subset~\citep{Li2025webthinker, wu2025webdancer}.
    \item \textbf{BrowseComp-en}~\citep{bc_en} - A benchmark of 1,266 questions probing agents' ability to locate and integrate hard-to-find, interrelated web information, with an emphasis on persistent browsing and factual reasoning.
    \item \textbf{BrowseComp-zh}~\citep{bc_zh} - A Chinese web-browsing benchmark with 289 multi-hop questions highlighting retrieval and reasoning challenges specific to the Chinese information ecosystem.

    \item \textbf{Xbench-DeepSearch}~\citep{xbench} - A specialized deep-search benchmark evaluating agents' end-to-end capabilities in planning, searching, reasoning, and summarization. Featuring expert-curated questions with broad search spaces and deep reasoning requirements, it complements existing benchmarks with substantial Chinese-context coverage.


    \item \textbf{FRAMES}~\citep{krishna2024fact} - A comprehensive RAG benchmark with 824 questions testing factuality, retrieval quality, and multi-hop reasoning. It evaluates models' ability to synthesize information from multiple retrieved sources while maintaining factual accuracy and reasoning coherence.

    
\end{itemize}

\paragraph{Baselines} We compare our \papername against the following baselines:
\begin{itemize}
    \item \textbf{General LLMs with Tools}: Models equipped with external tools for complex reasoning. We evaluate Qwen3-30B-A3B, Qwen3-235B-A22B~\citep{yang2025qwen3}, Claude-4-Sonnet~\citep{claude_deep_research}, OpenAI-o3~\citep{o3},  DeepSeek-V3.1 and DeepSeek-R1~\citep{r1}, GLM-4.5~\citep{zeng2025glm}, and Kimi-K2~\citep{team2025kimi}.
    
    \item \textbf{Commercial Deep Research Agents}: We test OpenAI's DeepResearch~\citep{dr}, Gemini Deep Research~\citep{google_dr}, Perplexity Deep Research~\citep{Perplexity}, Grok-DeepResearch~\citep{grok3}, and Kimi-Researcher~\citep{kimi-researcher}. However, as not all of them are fully accessible via API, they were not tested across all benchmarks and experiments.
    
    \item \textbf{Open-source Deep Research Agents}: We compare our method with recent open-source web/search agents, including WebDancer~\citep{wu2025webdancer}, WebSailor~\citep{li2025websailor}, MiroThinker~\citep{MiroThinker}, WebExplorer~\citep{liu2025webexplorer}. These represent the current state-of-the-art in open-source web research systems.
\end{itemize}

\paragraph{Tools} Our framework equips agents with four essential tools that enable comprehensive research capabilities, from information discovery to computational analysis. Each tool is designed to handle batch operations efficiently and return structured outputs suitable for iterative research processes.
\begin{itemize}
    \item \textbf{Search} enables web information retrieval via Google search engine. It accepts multiple queries simultaneously and returns top-10 results for each, with each result containing title, snippet, and URL for quick relevance assessment.
    \item \textbf{Scholar} provides access to academic literature through Google Scholar. Similar to Search, it supports batch queries and returns scholarly metadata including authors, venues, and citation counts, enabling efficient academic research.
    \item \textbf{Visit} extracts detailed content from specific web pages with goal-oriented summarization. The agent provides URLs along with extraction goals (e.g., "find experimental results"), and the tool first retrieves full content via Jina~\citep{jina}, then uses Qwen3~\citep{yang2025qwen3} to produce focused summaries based on the specified goals.
    \item \textbf{Python} executes code in a sandboxed environment for computational tasks. It supports standard libraries for data analysis and visualization, with all outputs explicitly printed to ensure clear result communication.
\end{itemize}

\paragraph{Evaluation Metrics and Hyper-parameters}
We adopt the \textbf{pass@$k$} metric~\citep{chen2021evaluating} to evaluate the model's performance. In our experiments, we primarily report \textbf{pass@1}, which represents the percentage of problems solved correctly in a single attempt. To determine the correctness of a generated solution, we employ an \textbf{LLM-as-a-Judge} approach~\citep{DBLP:conf/coling/LiuYHZHWDSZ24, DBLP:conf/emnlp/WangCCL0WYXZLLY24}. For all generation tasks, we use nucleus sampling with a \textbf{temperature} of 0.6 and a \textbf{top-p} of 0.95.

For a dataset with $n$ problems, pass@1 is formally calculated as:
\begin{equation}
\label{eq:pass_at_1}
\text{pass@1} = \frac{1}{n} \sum_{i=1}^n \mathbb{I}(\text{problem } i \text{ is solved}),
\end{equation}
where $\mathbb{I}(\cdot)$ is the indicator function. For pass@$k$ where $k>1$, we generate $k$ independent samples per problem and consider it solved if at least one sample is correct.

\subsection{Main Results}
\begin{table}[h]
    \caption{Results on General Web Navigation and Reasoning Benchmarks. $^\dagger$ marks the result from the corresponding official reports.}
    \centering
    \resizebox{\textwidth}{!}{
    \begin{tabular}{l|c|c|c}
    \toprule
    \textbf{Backbone} & \textbf{Humanity's Last Exam} & \textbf{BrowseComp} & \textbf{BrowseComp-ZH} \\
    \midrule
    \rowcolor[RGB]{229,229,252}\multicolumn{4}{c}{\emph{\textbf{General LLMs with tools}}} \\
    \midrule
    Qwen3-30B-A3B & 13.2 & 0.5 & 13.5 \\
    Qwen3-235B-A22B & 20.0 & 2.3 & 29.4 \\
    DeepSeek-R1 & 24.8$^\dagger$ & 8.9$^\dagger$ & 35.7$^\dagger$ \\
    Claude-4-Sonnet & 20.3$^\dagger$ & 12.2$^\dagger$ & 29.1$^\dagger$ \\
    \midrule
    \rowcolor[RGB]{229,229,252}\multicolumn{4}{c}{\emph{\textbf{Commercial Deep Research Agents}}} \\
    \midrule
    Perplexity Deep Research & 21.1$^\dagger$ & - & - \\
    Gemini Deep Research & 26.9$^\dagger$ & - & - \\
    Kimi-Researcher & 26.9$^\dagger$ & - & - \\
    OpenAI-o3 & 20.2$^\dagger$ & 49.7$^\dagger$ & 58.1$^\dagger$ \\
    OpenAI Deep Research & 26.6$^\dagger$ & 51.5$^\dagger$ & - \\
    \midrule
    \rowcolor[RGB]{229,229,252}\multicolumn{4}{c}{\emph{\textbf{Open-source Deep Research Agents}}} \\
    \midrule
    WebSailor-72B & - & 12.0$^\dagger$ & 30.1$^\dagger$ \\
    WebShaper-72B & - & - & - \\
    MiroThinker-32B & - & 13.0$^\dagger$ & 17.0$^\dagger$ \\
    WebExplorer-8B & - & 15.7$^\dagger$ & 32.0$^\dagger$ \\
    Kimi-K2 & 18.1$^\dagger$ & 14.1$^\dagger$ & 28.8$^\dagger$ \\
    GLM-4.5 & 21.2$^\dagger$ & 26.4$^\dagger$ & 37.5$^\dagger$ \\
    DeepSeek-V3.1 & 29.8$^\dagger$ & 30.0$^\dagger$ & 49.2$^\dagger$ \\
    \midrule
    \rowcolor[RGB]{229,229,252}\multicolumn{4}{c}{\emph{\textbf{Ours}}} \\
    \midrule
    \textbf{\papername-30B-A3B} & \textbf{28.8} & \textbf{37.3} & \textbf{45.2} \\
    \textbf{\papername-30B-A3B-heavy} & \textbf{36.7} & \textbf{51.7} & \textbf{56.8} \\
    \bottomrule
    \end{tabular}
    }
    \label{tab:scenario-targeted-web-search}
\end{table}

\begin{table}[h]
    \caption{Results on Complex, Goal-Oriented Web Tasks Benchmarks. $^\dagger$ marks the result from the corresponding official reports.}
    \centering
    \resizebox{0.8\textwidth}{!}{
    \begin{tabular}{l|c|c|c}
    \toprule
    \textbf{Backbone} & \textbf{GAIA} & \textbf{Xbench-DeepSearch} & \textbf{Frames} \\
    \midrule
    \rowcolor[RGB]{229,229,252}\multicolumn{4}{c}{\emph{\textbf{General LLMs with tools}}} \\
    \midrule
    Qwen3-30B-A3B & 35.9 & 32.0 & 56.4 \\
    Qwen3-235B-A22B & 45.6 & 46.0 & - \\
    DeepSeek-R1 & - & 55.0$^\dagger$ & 82.0$^\dagger$ \\
    Claude-4-Sonnet & 68.3$^\dagger$ & 64.6$^\dagger$ & 80.7$^\dagger$ \\
    \midrule
    \rowcolor[RGB]{229,229,252}\multicolumn{4}{c}{\emph{\textbf{Commercial Deep Research Agents}}} \\
    \midrule
    Kimi-Researcher & - & 69.0$^\dagger$ & 78.8$^\dagger$ \\
    OpenAI-o3 & 70.5$^\dagger$ & 66.7$^\dagger$ & 84.0$^\dagger$ \\
    OpenAI Deep Research & 67.0$^\dagger$ & - & - \\
    \midrule
    \rowcolor[RGB]{229,229,252}\multicolumn{4}{c}{\emph{\textbf{Open-source Deep Research Agents}}} \\
    \midrule
    WebSailor-72B & - & 55.0$^\dagger$ & - \\
    WebExplorer-8B & - & 53.7$^\dagger$ & - \\
    Kimi-K2 & 57.3$^\dagger$ & 50.0$^\dagger$ & 72.0$^\dagger$ \\
    GLM-4.5 & 66.0$^\dagger$ & 70.0$^\dagger$ & 78.9$^\dagger$ \\
    DeepSeek-V3.1 & 63.1$^\dagger$ & 71.2$^\dagger$ & 83.7$^\dagger$ \\
    \midrule
    \rowcolor[RGB]{229,229,252}\multicolumn{4}{c}{\emph{\textbf{Ours}}} \\
    \midrule
    \textbf{\papername-30B-A3B} & \textbf{72.8} & \textbf{71.0} & \textbf{84.8} \\
    \textbf{\papername-30B-A3B-heavy} & \textbf{75.7} & \textbf{73.0} & \textbf{85.1} \\
    \bottomrule
    \end{tabular}
    }
    \label{tab:general-web-search}
\end{table}

We present comprehensive evaluation results across 6 challenging benchmarks, categorized into complex goal-oriented web tasks (Table~\ref{tab:general-web-search}) and general web navigation and reasoning challenges (Table~\ref{tab:scenario-targeted-web-search}).

\paragraph{Overall Performance.}
\papername demonstrates state-of-the-art performance across diverse deep-research benchmarks, significantly outperforming both larger models and existing deep-research systems. \papername achieves remarkable results that surpass open-source deep-research agents and even proprietary deep-research systems. Across 6 challenging benchmarks spanning complex reasoning, web navigation, and long-horizon information-seeking tasks, \papername consistently ranks among the top performers, validating the effectiveness of our iterative synthesis paradigm over the prevalent mono-contextual approach.

\paragraph{General Web Navigation and Reasoning Benchmarks.}
Table~\ref{tab:scenario-targeted-web-search} demonstrates \papername's superior performance on web-scale information synthesis tasks, where the advantages of our iterative paradigm become most pronounced. On Humanity's Last Exam (HLE), arguably one of the most challenging benchmarks for frontier AI systems, \papername-heavy achieves 36.7\% accuracy—dramatically outperforming all systems including DeepSeek-V3.1 (29.8\%), OpenAI Deep Research (26.6\%), and Gemini Deep Research (26.9\%). This exceptional 6.9 percentage point improvement over the next best system on HLE, which requires deep academic knowledge synthesis across multiple disciplines, validates our paradigm's core strength: maintaining deep reasoning capabilities throughout extended research processes by ensuring each round operates with full cognitive capacity rather than diminishing workspace.

The performance gains are equally impressive on web navigation benchmarks. On BrowseComp-en, \papername-heavy achieves 51.7\% accuracy, matching OpenAI's Deep Research (51.5\%) while vastly exceeding all open-source alternatives—DeepSeek-V3.1, the next best open-source system, achieves only 30.0\%. This 21.7 percentage point improvement demonstrates the critical importance of our iterative synthesis approach when handling complex web navigation tasks that require maintaining coherent understanding across multiple information sources.

Similarly strong results are observed on the Chinese-language benchmark BrowseComp-zh, where \papername-heavy achieves 56.8\%, approaching o3's 58.1\% while significantly outperforming DeepSeek-V3.1 (49.2\%). These multilingual results highlight that our iterative paradigm effectively handles culturally-diverse information sources through its structured synthesis process—each round's report distills cross-lingual insights into a coherent narrative, preventing the confusion that often arises when mono-contextual systems accumulate mixed-language content without proper consolidation mechanisms.

\paragraph{Complex Goal-Oriented Web Tasks.}
Table~\ref{tab:general-web-search} reveals \papername's exceptional capability in handling complex, multi-step reasoning tasks. On GAIA, \papername achieves 72.8\% accuracy, surpassing all evaluated systems including Claude-4-Sonnet (68.3\%) and OpenAI-o3 (70.5\%), with a remarkable 9.7 percentage point improvement over DeepSeek-V3.1 (63.1\%). This substantial gain demonstrates the superiority of iterative synthesis when tackling tasks that require sophisticated tool orchestration and cross-domain reasoning. The key advantage of our iterative paradigm becomes evident here: by periodically reconstructing the workspace and synthesizing findings, \papername maintains consistent reasoning quality throughout extended research processes, whereas mono-contextual systems suffer from progressive degradation due to context bloat.

On Xbench-DeepSearch, our system reaches 71.0\%, matching DeepSeek-V3.1 (71.2\%) while vastly exceeding other open-source alternatives like WebSailor-72B (55.0\%) and Kimi-K2 (50.0\%). Similarly impressive results are observed on Frames (84.8\%), where \papername outperforms all systems including DeepSeek-V3.1 (83.7\%) and OpenAI-o3 (84.0\%). These consistent improvements across diverse task types reveal the fundamental advantage of iterative synthesis: by periodically consolidating findings and reconstructing focused workspaces, \papername can pursue complex reasoning chains and adapt search strategies based on synthesized insights—capabilities that mono-contextual systems inherently lack due to their linear accumulation constraints.
\section{Analysis}

\subsection{The Primacy of the Iterative Paradigm}
To verify that the performance gains of our model stem from its core design rather than confounding factors, we conducted a targeted ablation study. The objective was to isolate and measure the direct impact of our Iterative Deep-Research Paradigm.
\paragraph{Experimental Setup} We designed an ablation variant, herein referred to as Mono-Agent. This agent utilizes the same underlying model architecture as our full agent but is constrained to a linear, non-iterative inference strategy. Specifically, it accumulates all generated information—including thoughts, tool interactions, and observations—into a single, continuously expanding context window, lacking any mechanism for synthesis or reset. We compare this against two other agents: Mono-Agent + Iter, which represents the Mono-Agent architecture enhanced with our iterative research training data but still using the linear inference strategy, and WebResearcher, our full model employing the iterative paradigm.

\begin{wraptable}{r}{0.5\textwidth}
    \vspace{-15pt} 
    \centering
    \caption{Main results comparing different agents on the HLE, BC-EN, and BC-ZH benchmarks.}
    \label{tab:agent_performance}
    \begin{tabular}{@{}lccc@{}}
        \toprule
        \textbf{Agent} & \textbf{HLE} & \textbf{BC-EN} & \textbf{BC-ZH} \\
        \midrule
        WebResearcher      & \textbf{28.8} & \textbf{37.3} & \textbf{45.2} \\
        Mono-Agent        & 18.7 & 25.4 & 34.6 \\
        Mono-Agent + Iter & 25.4 & 30.1 & 40.4 \\
        \bottomrule
    \end{tabular}
\end{wraptable}

\paragraph{Results and Analysis} The results, presented in Table~\ref{tab:agent_performance}, clearly demonstrate the paradigm's efficacy. The Mono-Agent + Iter consistently outperforms the base Mono-Agent across all benchmarks: HLE (25.4 vs. 18.7), BC-EN (30.1 vs. 25.4), and BC-ZH (40.4 vs. 34.6). This initial improvement highlights the benefit of our specialized training data.

However, the most significant finding is the performance gap between the non-iterative Mono-Agent + Iter and our full WebResearcher agent (e.g., 28.8 vs. 25.4 on HLE). This delta isolates the impact of the iterative paradigm itself. The inferior performance of the linear strategy is attributable to two critical failure modes: 1) Contextual Degradation, where the model's attention is saturated with an excess of low-value historical data, impairing its ability to identify salient information; and 2) Irreversible Error Propagation, where early mistakes or noisy observations remain in the context, progressively corrupting subsequent reasoning steps. This is particularly detrimental in long-horizon tasks that require numerous steps.

Conversely, our iterative paradigm directly mitigates these issues. By periodically synthesizing key findings and resetting the contextual workspace, our agent maintains a focused and refined context for each reasoning cycle. This mechanism is fundamental to sustaining high-level cognitive performance. This study provides compelling evidence that the iterative paradigm itself, not merely the training data or base model, is the critical driver of WebResearcher's success in complex, long-horizon research tasks.

\subsection{Analysis of Tool-Use Behavior}
The core strength of \modelname lies in its iterative paradigm, which facilitates longer and more complex reasoning chains. To substantiate this claim, we conduct an in-depth analysis of its tool-calling behavior across diverse benchmarks, demonstrating that \modelname exhibits highly adaptive and efficient tool-use strategies tailored to task-specific demands.

Our analysis focuses on the frequency and length of tool invocation sequences—specifically involving Search (web search), Scholar (academic search), Visit (web page access), and Python (code execution)—on the HLE and BrowseComp benchmarks. The tool-use profile of \modelname shifts dramatically based on the nature of the task.

On the HLE benchmark, which primarily contains questions requiring academic and professional knowledge, the agent adopts a focused and concise strategy. The Scholar tool is prominently used, constituting 25.4\% of all tool calls, reflecting the need for specialized literature search. The average reasoning chain is short, with tasks being resolved in an average of only 4.7 turns. This indicates efficient, targeted information retrieval for well-defined problems.

In stark contrast, on BrowseComp, where tasks necessitate extensive web navigation and information integration across multiple pages, the agent's behavior highlights its capacity for prolonged and complex reasoning. The Search (56.5\%) and Visit (39.7\%) tools become paramount, jointly accounting for over 96\% of all tool invocations. This strategic shift is mirrored by a significant increase in reasoning complexity: the average number of turns skyrockets to 61.4 per task, with the most complex problems requiring over 200 interaction turns to solve.

This marked divergence in both tool selection and reasoning chain length underscores \modelname's sophisticated ability to dynamically adapt its problem-solving approach. It can execute brief, precise actions for knowledge-based queries (HLE) as well as sustain long, exploratory sequences for complex web-based tasks (BrowseComp), validating the effectiveness of its iterative reasoning architecture.

\begin{figure}[t]
    \centering
    \begin{subfigure}{0.48\textwidth}
        \centering
        \includegraphics[width=\linewidth]{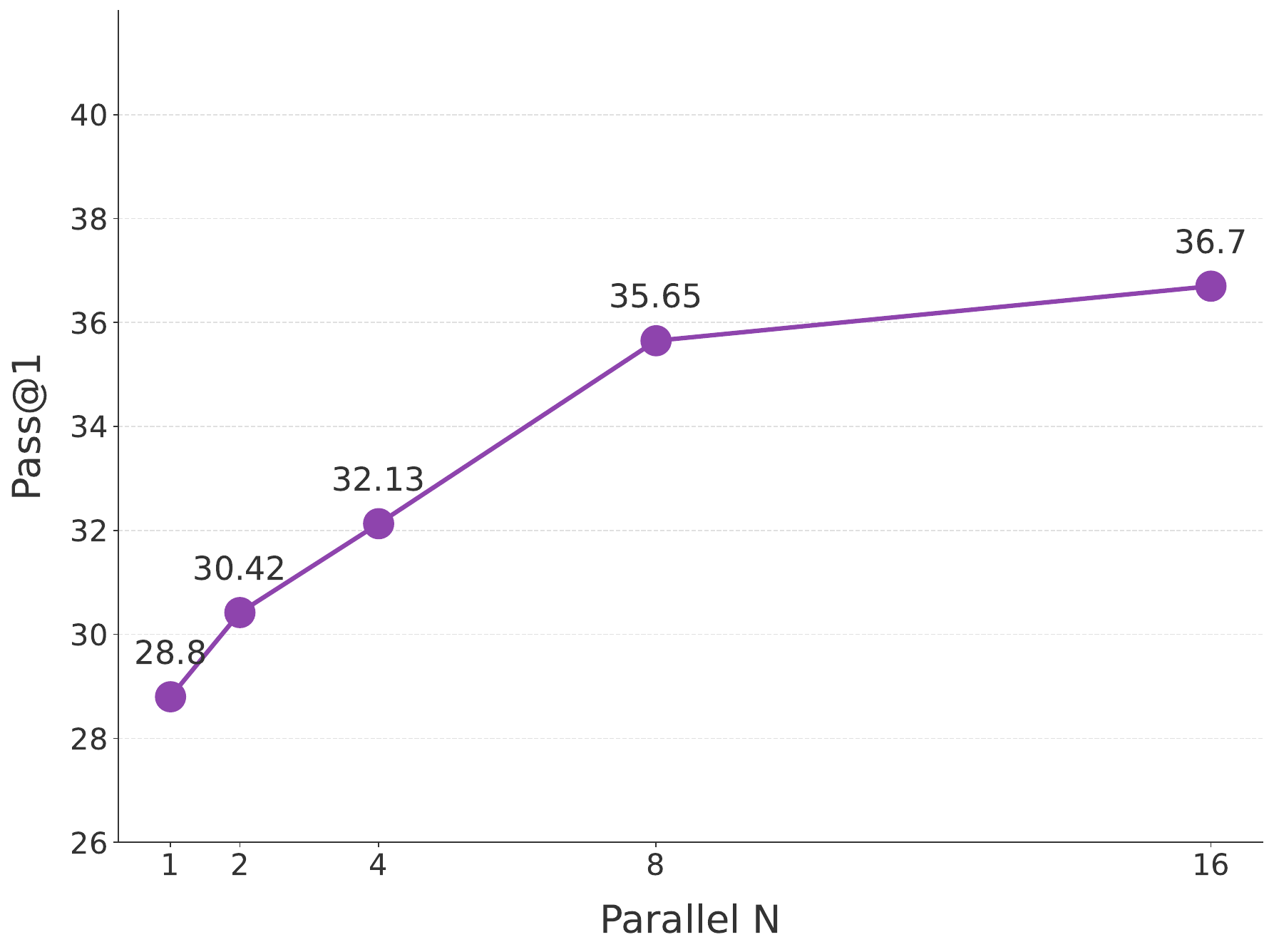} 
        \caption{Effect of n on HLE}
        \label{fig:hle-tts}
    \end{subfigure}
    \hfill
    \begin{subfigure}{0.48\textwidth}
        \centering
        \includegraphics[width=\linewidth]{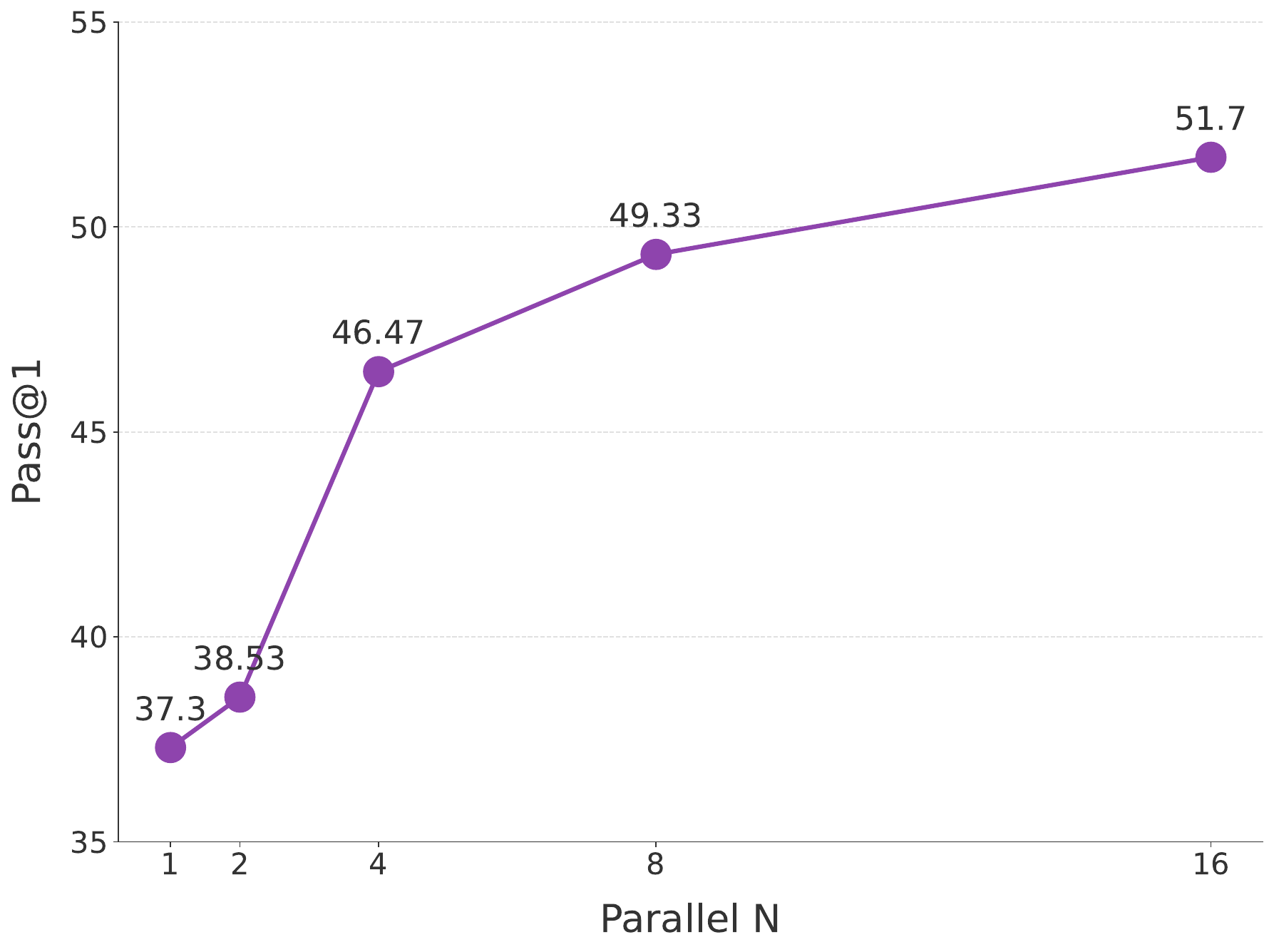}
        \caption{Effect of n on BrowseComp}
        \label{fig:bc_en}
    \end{subfigure}
    \caption{Effect of n in Reason-Synthesis Framework}
    \label{fig:tts_n_analysis}
\end{figure}

\subsection{Analysis on Reasoning Trajectories in Reason-Synthesis Framework}
In Section~\ref{sec:tts}, we introduced Reason-synthesis Framework to enhance model performance by running $n$ inference trajectories in parallel. To consolidate the findings from all Research Agents, a Synthesis Agent aggregates these reasoning paths to generate the final answer. This section presents a quantitative analysis of the impact of the hyperparameter $n$—the number of parallel research—on final performance.

\paragraph{Experimental Setup}
Our analysis is conducted on the HLE benchmark using the \modelname-30B-A3B model. We systematically vary the number of parallel research, $n$, evaluating the model's performance (pass@1) for $n \in \{1, 2, 4, 8, 16\}$. The case of $n=1$ serves as the baseline, representing the model's performance without any Test-Time Scaling.

\paragraph{Results and Insights}
The experimental results, illustrated in Figure \ref{fig:tts_n_analysis}, reveal a clear and positive correlation between the number of trajectories ($n$) and model performance. 

As we increase $n$, there is a consistent and significant improvement in the pass@1 score. The most substantial gains are observed when scaling $n$ from 1 to 8. This result underscores the benefits of the \textbf{ensemble effect} inherent in TTS. Each Research Agent explores a unique reasoning path, potentially uncovering distinct facets of the problem or overcoming specific intermediate obstacles. By fusing the final outcomes of these diverse explorations, the Synthesis Agent can produce a more robust and accurate final answer.

As expected, these performance improvements are accompanied by a linear increase in computational cost, as each trajectory is processed independently. Furthermore, the performance gains begin to exhibit \textbf{diminishing marginal returns} for $n > 8$, indicating a trade-off between accuracy and computational budget.

Our analysis demonstrates that employing multiple parallel research trajectories within the Reason-Synthesis framework is a highly effective technique for performance enhancement. The number of trajectories, $n$, serves as a direct and controllable parameter to balance performance gains against computational cost. Based on our findings, a configuration of $n=8$ offers a compelling trade-off, delivering substantial performance improvements over the baseline while maintaining manageable computational overhead.


\section{Related Work}
\paragraph{Deep Research}
The development of autonomous deep-research agents has witnessed significant progress from both proprietary and open-source efforts.
These proprietary systems~\citep{dr,google_dr,grok3,claude_deep_research,Perplexity,kimi-researcher} have established benchmarks for deep-research capabilities but remain opaque.
In contrast, open-source efforts~\citep{chen2025cpo,jin2025search,li2025search,Li2025webthinker,tao2025webshaper,li2025websailor,agentscaler,agentfounder2025} have advanced the field through transparent architectures and training methodologies. However, these works predominantly adopt the mono-contextual paradigm, accumulating all retrieved information into a single, ever-expanding context. While this linear approach has shown initial success, it fundamentally constrains reasoning capacity as contexts bloat and allows noise contamination to persist throughout the research process.
\papername departs from this prevalent paradigm by introducing an iterative synthesis framework that maintains focused cognitive workspaces through periodic consolidation and reconstruction. Our approach draws inspiration from human research workflows~\citep{bellman1957markovian,puterman1990markov}, where researchers iteratively refine their understanding through cycles of exploration, synthesis, and focused investigation. Unlike mono-contextual systems that suffer from irreversible degradation, \papername sustains high-quality reasoning at arbitrary research depths through its Markov Decision Process formulation, enabling complex multi-hop reasoning and cross-domain synthesis that existing architectures struggle to achieve.
\section{Conclusion}
In this paper, we presented \papername, a novel framework that fundamentally rethinks deep-research agents through three key contributions: (1) \modelname, an iterative paradigm that reformulates deep research as a Markov Decision Process with periodic consolidation, overcoming the context suffocation and noise contamination of mono-contextual approaches; (2) \dataname, a scalable data synthesis engine that addresses training data scarcity through tool-augmented complexity escalation; and (3) a Research-Synthesis Framework that enables effective test-time scaling through parallel multi-agent exploration. Extensive experiments across 6 challenging benchmarks demonstrate that \papername achieves state-of-the-art performance, surpassing even frontier proprietary systems. These results validate our core insight that effective deep research requires structured iteration with periodic synthesis rather than unbounded accumulation.


\clearpage
\bibliography{biblio}

\begin{thebibliography}{41}
\providecommand{\natexlab}[1]{#1}
\providecommand{\url}[1]{\texttt{#1}}
\expandafter\ifx\csname urlstyle\endcsname\relax
  \providecommand{\doi}[1]{doi: #1}\else
  \providecommand{\doi}{doi: \begingroup \urlstyle{rm}\Url}\fi

\bibitem[Anthropic(2024)]{claude}
Anthropic.
\newblock Introducing computer use, a new claude 3.5 sonnet, and claude 3.5 haiku.
\newblock October 2024.

\bibitem[Anthropic(2025)]{claude_deep_research}
Anthropic.
\newblock Claude takes research to new places.
\newblock \url{https://www.anthropic.com/news/research}, April 2025.

\bibitem[Bellman(1957)]{bellman1957markovian}
Richard Bellman.
\newblock A markovian decision process.
\newblock \emph{Journal of mathematics and mechanics}, pp.\  679--684, 1957.

\bibitem[Chen et~al.(2025)Chen, Liao, Yu, Wang, Qiao, Yang, Zhao, and Fan]{chen2025cpo}
Guoxin Chen, Minpeng Liao, Peiying Yu, Dingmin Wang, Zile Qiao, Chao Yang, Xin Zhao, and Kai Fan.
\newblock C-3{PO}: Compact plug-and-play proxy optimization to achieve human-like retrieval-augmented generation.
\newblock In \emph{Forty-second International Conference on Machine Learning}, 2025.
\newblock URL \url{https://openreview.net/forum?id=hlpwAmQ4wr}.

\bibitem[Chen et~al.(2021)Chen, Tworek, Jun, Yuan, Pinto, Kaplan, Edwards, Burda, Joseph, Brockman, et~al.]{chen2021evaluating}
Mark Chen, Jerry Tworek, Heewoo Jun, Qiming Yuan, Henrique Ponde De~Oliveira Pinto, Jared Kaplan, Harri Edwards, Yuri Burda, Nicholas Joseph, Greg Brockman, et~al.
\newblock Evaluating large language models trained on code.
\newblock \emph{arXiv preprint arXiv:2107.03374}, 2021.

\bibitem[Fang et~al.(2025)Fang, Cai, Li, Wu, Li, Yin, Wang, Wang, Su, Zhang, Wu, Tao, Jiang, Xie, Huang, and Zhou]{agentscaler}
Runnan Fang, Shihao Cai, Baixuan Li, Jialong Wu, Guangyu Li, Wenbiao Yin, Xinyu Wang, Xiaobin Wang, Liangcai Su, Zhen Zhang, Shibin Wu, Zhengwei Tao, Yong Jiang, Pengjun Xie, Fei Huang, and Jingren Zhou.
\newblock Towards general agentic intelligence via environment scaling, 2025.

\bibitem[Gemma et~al.(2025)Gemma, Kamath, Ferret, Pathak, Vieillard, Merhej, Perrin, Matejovicova, Ram{\'e}, Rivi{\`e}re, et~al.]{gemma3}
Team Gemma, Aishwarya Kamath, Johan Ferret, Shreya Pathak, Nino Vieillard, Ramona Merhej, Sarah Perrin, Tatiana Matejovicova, Alexandre Ram{\'e}, Morgane Rivi{\`e}re, et~al.
\newblock Gemma 3 technical report.
\newblock \emph{arXiv preprint arXiv:2503.19786}, 2025.

\bibitem[Google(2025{\natexlab{a}})]{gemini_25_pro}
Google.
\newblock Gemini 2.5 pro.
\newblock \url{https://deepmind.google/technologies/gemini/pro/}, April 2025{\natexlab{a}}.

\bibitem[Google(2025{\natexlab{b}})]{google_dr}
Google.
\newblock Deep research is now available on gemini 2.5 pro experimental., 2025{\natexlab{b}}.
\newblock URL \url{https://blog.google/products/gemini/deep-research-gemini-2-5-pro-experimental/}.

\bibitem[Guo et~al.(2025{\natexlab{a}})Guo, Yang, Zhang, Song, Zhang, Xu, Zhu, Ma, Wang, Bi, et~al.]{guo2025deepseek}
Daya Guo, Dejian Yang, Haowei Zhang, Junxiao Song, Ruoyu Zhang, Runxin Xu, Qihao Zhu, Shirong Ma, Peiyi Wang, Xiao Bi, et~al.
\newblock Deepseek-r1: Incentivizing reasoning capability in llms via reinforcement learning.
\newblock \emph{arXiv preprint arXiv:2501.12948}, 2025{\natexlab{a}}.

\bibitem[Guo et~al.(2025{\natexlab{b}})Guo, Yang, Zhang, Song, Zhang, Xu, Zhu, Ma, Wang, Bi, et~al.]{r1}
Daya Guo, Dejian Yang, Haowei Zhang, Junxiao Song, Ruoyu Zhang, Runxin Xu, Qihao Zhu, Shirong Ma, Peiyi Wang, Xiao Bi, et~al.
\newblock {DeepSeek-R1}: Incentivizing reasoning capability in {LLMs} via reinforcement learning.
\newblock \emph{arXiv preprint arXiv:2501.12948}, 2025{\natexlab{b}}.

\bibitem[Jin et~al.(2025)Jin, Zeng, Yue, Yoon, Arik, Wang, Zamani, and Han]{jin2025search}
Bowen Jin, Hansi Zeng, Zhenrui Yue, Jinsung Yoon, Sercan Arik, Dong Wang, Hamed Zamani, and Jiawei Han.
\newblock Search-r1: Training llms to reason and leverage search engines with reinforcement learning.
\newblock \emph{arXiv preprint arXiv:2503.09516}, 2025.

\bibitem[Jina.ai(2025)]{jina}
Jina.ai.
\newblock Jina, 2025.
\newblock URL \url{https://jina.ai/}.

\bibitem[Krishna et~al.(2024)Krishna, Krishna, Mohananey, Schwarcz, Stambler, Upadhyay, and Faruqui]{krishna2024fact}
Satyapriya Krishna, Kalpesh Krishna, Anhad Mohananey, Steven Schwarcz, Adam Stambler, Shyam Upadhyay, and Manaal Faruqui.
\newblock Fact, fetch, and reason: A unified evaluation of retrieval-augmented generation.
\newblock \emph{arXiv preprint arXiv:2409.12941}, 2024.

\bibitem[Li et~al.(2025{\natexlab{a}})Li, Zhang, Yin, Zhang, Ou, Wu, Yin, Li, Tao, Wang, et~al.]{li2025websailor}
Kuan Li, Zhongwang Zhang, Huifeng Yin, Liwen Zhang, Litu Ou, Jialong Wu, Wenbiao Yin, Baixuan Li, Zhengwei Tao, Xinyu Wang, et~al.
\newblock Websailor: Navigating super-human reasoning for web agent.
\newblock \emph{arXiv preprint arXiv:2507.02592}, 2025{\natexlab{a}}.

\bibitem[Li et~al.(2025{\natexlab{b}})Li, Dong, Jin, Zhang, Zhou, Zhu, Zhang, and Dou]{li2025search}
Xiaoxi Li, Guanting Dong, Jiajie Jin, Yuyao Zhang, Yujia Zhou, Yutao Zhu, Peitian Zhang, and Zhicheng Dou.
\newblock Search-o1: Agentic search-enhanced large reasoning models.
\newblock \emph{arXiv preprint arXiv:2501.05366}, 2025{\natexlab{b}}.

\bibitem[Li et~al.(2025{\natexlab{c}})Li, Jin, Dong, Qian, Zhu, Wu, Wen, and Dou]{Li2025webthinker}
Xiaoxi Li, Jiajie Jin, Guanting Dong, Hongjin Qian, Yutao Zhu, Yongkang Wu, Ji{-}Rong Wen, and Zhicheng Dou.
\newblock Webthinker: Empowering large reasoning models with deep research capability.
\newblock \emph{CoRR}, abs/2504.21776, 2025{\natexlab{c}}.
\newblock \doi{10.48550/ARXIV.2504.21776}.
\newblock URL \url{https://doi.org/10.48550/arXiv.2504.21776}.

\bibitem[Liu et~al.(2025)Liu, Li, Zhang, Li, Chen, Ji, Cheng, Wu, Du, Xu, et~al.]{liu2025webexplorer}
Junteng Liu, Yunji Li, Chi Zhang, Jingyang Li, Aili Chen, Ke~Ji, Weiyu Cheng, Zijia Wu, Chengyu Du, Qidi Xu, et~al.
\newblock Webexplorer: Explore and evolve for training long-horizon web agents.
\newblock \emph{arXiv preprint arXiv:2509.06501}, 2025.

\bibitem[Liu et~al.(2024)Liu, Yang, Huang, Zhang, Huang, Wei, Deng, Sun, and Zhang]{DBLP:conf/coling/LiuYHZHWDSZ24}
Yuxuan Liu, Tianchi Yang, Shaohan Huang, Zihan Zhang, Haizhen Huang, Furu Wei, Weiwei Deng, Feng Sun, and Qi~Zhang.
\newblock Calibrating llm-based evaluator.
\newblock In Nicoletta Calzolari, Min{-}Yen Kan, V{\'{e}}ronique Hoste, Alessandro Lenci, Sakriani Sakti, and Nianwen Xue (eds.), \emph{Proceedings of the 2024 Joint International Conference on Computational Linguistics, Language Resources and Evaluation, {LREC/COLING} 2024, 20-25 May, 2024, Torino, Italy}, pp.\  2638--2656. {ELRA} and {ICCL}, 2024.
\newblock URL \url{https://aclanthology.org/2024.lrec-main.237}.

\bibitem[Meta(2025)]{llama4}
Team Meta.
\newblock The llama 4 herd: The beginning of a new era of natively multimodal ai innovation.
\newblock \url{https://ai.meta.com/blog/llama-4-multimodal-intelligence/}, April 2025.

\bibitem[Mialon et~al.(2023)Mialon, Fourrier, Wolf, LeCun, and Scialom]{mialon2023gaia}
Gr{\'e}goire Mialon, Cl{\'e}mentine Fourrier, Thomas Wolf, Yann LeCun, and Thomas Scialom.
\newblock Gaia: a benchmark for general ai assistants.
\newblock In \emph{The Twelfth International Conference on Learning Representations}, 2023.

\bibitem[MiroMindAI(2025)]{MiroThinker}
MiroMindAI.
\newblock Mirothinker, 2025.
\newblock URL \url{https://github.com/MiroMindAI/MiroThinker}.

\bibitem[MoonshotAI(2025)]{kimi-researcher}
MoonshotAI.
\newblock Kimi-researcher, 2025.
\newblock URL \url{https://moonshotai.github.io/Kimi-Researcher/}.

\bibitem[OpenAI(2025{\natexlab{a}})]{dr}
OpenAI.
\newblock Deep research system card, 2025{\natexlab{a}}.
\newblock URL \url{https://cdn.openai.com/deep-research-system-card.pdf}.

\bibitem[OpenAI(2025{\natexlab{b}})]{o3}
OpenAI.
\newblock Introducing openai o3 and o4-mini, 2025{\natexlab{b}}.
\newblock URL \url{https://openai.com/index/introducing-o3-and-o4-mini/}.

\bibitem[OpenAI(2025{\natexlab{c}})]{openai_o3_o4_mini}
OpenAI.
\newblock Introducing openai o3 and o4-mini.
\newblock \url{https://openai.com/index/introducing-o3-and-o4-mini/}, April 2025{\natexlab{c}}.

\bibitem[Perplexity(2025)]{Perplexity}
Perplexity.
\newblock Introducing perplexity deep research, 2025.
\newblock URL \url{https://www.perplexity.ai/hub/blog/introducing-perplexity-deep-research}.

\bibitem[Phan et~al.(2025)Phan, Gatti, Han, Li, Hu, Zhang, Zhang, Shaaban, Ling, Shi, et~al.]{hle}
Long Phan, Alice Gatti, Ziwen Han, Nathaniel Li, Josephina Hu, Hugh Zhang, Chen Bo~Calvin Zhang, Mohamed Shaaban, John Ling, Sean Shi, et~al.
\newblock Humanity's last exam.
\newblock \emph{arXiv preprint arXiv:2501.14249}, 2025.

\bibitem[Puterman(1990)]{puterman1990markov}
Martin~L Puterman.
\newblock Markov decision processes.
\newblock \emph{Handbooks in operations research and management science}, 2:\penalty0 331--434, 1990.

\bibitem[Su et~al.(2025)Su, Zhang, Li, Chen, Wang, Song, Wang, Li, Wu, Chen, Qiao, Zhang, Yin, Cai, Fang, Tao, Yin, et~al.]{agentfounder2025}
Liangcai Su, Zhen Zhang, Guangyu Li, Zhuo Chen, Chenxi Wang, Maojia Song, Xinyu Wang, Kuan Li, Jialong Wu, Xuanzhong Chen, Zile Qiao, Zhongwang Zhang, Huifeng Yin, Shihao Cai, Runnan Fang, Zhengwei Tao, Wenbiao Yin, et~al.
\newblock Scaling agents via continual pre-training, 2025.

\bibitem[Tao et~al.(2025)Tao, Wu, Yin, Zhang, Li, Shen, Li, Zhang, Wang, Jiang, et~al.]{tao2025webshaper}
Zhengwei Tao, Jialong Wu, Wenbiao Yin, Junkai Zhang, Baixuan Li, Haiyang Shen, Kuan Li, Liwen Zhang, Xinyu Wang, Yong Jiang, et~al.
\newblock Webshaper: Agentically data synthesizing via information-seeking formalization.
\newblock \emph{arXiv preprint arXiv:2507.15061}, 2025.

\bibitem[Team et~al.(2025)Team, Bai, Bao, Chen, Chen, Chen, Chen, Chen, Chen, Chen, et~al.]{team2025kimi}
Kimi Team, Yifan Bai, Yiping Bao, Guanduo Chen, Jiahao Chen, Ningxin Chen, Ruijue Chen, Yanru Chen, Yuankun Chen, Yutian Chen, et~al.
\newblock Kimi k2: Open agentic intelligence.
\newblock \emph{arXiv preprint arXiv:2507.20534}, 2025.

\bibitem[Wang et~al.(2024)Wang, Chen, Cheng, Liao, Zhang, Wu, Yu, Xu, Zhang, Luo, Li, Yang, Huang, and Li]{DBLP:conf/emnlp/WangCCL0WYXZLLY24}
Minzheng Wang, Longze Chen, Fu~Cheng, Shengyi Liao, Xinghua Zhang, Bingli Wu, Haiyang Yu, Nan Xu, Lei Zhang, Run Luo, Yunshui Li, Min Yang, Fei Huang, and Yongbin Li.
\newblock Leave no document behind: Benchmarking long-context llms with extended multi-doc {QA}.
\newblock In Yaser Al{-}Onaizan, Mohit Bansal, and Yun{-}Nung Chen (eds.), \emph{Proceedings of the 2024 Conference on Empirical Methods in Natural Language Processing, {EMNLP} 2024, Miami, FL, USA, November 12-16, 2024}, pp.\  5627--5646. Association for Computational Linguistics, 2024.
\newblock URL \url{https://aclanthology.org/2024.emnlp-main.322}.

\bibitem[Wei et~al.(2025)Wei, Sun, Papay, McKinney, Han, Fulford, Chung, Passos, Fedus, and Glaese]{bc_en}
Jason Wei, Zhiqing Sun, Spencer Papay, Scott McKinney, Jeffrey Han, Isa Fulford, Hyung~Won Chung, Alex~Tachard Passos, William Fedus, and Amelia Glaese.
\newblock Browsecomp: A simple yet challenging benchmark for browsing agents.
\newblock \emph{arXiv preprint arXiv:2504.12516}, 2025.

\bibitem[Wu et~al.(2025)Wu, Li, Fang, Yin, Zhang, Tao, Zhang, Xi, Fu, Jiang, et~al.]{wu2025webdancer}
Jialong Wu, Baixuan Li, Runnan Fang, Wenbiao Yin, Liwen Zhang, Zhengwei Tao, Dingchu Zhang, Zekun Xi, Gang Fu, Yong Jiang, et~al.
\newblock Webdancer: Towards autonomous information seeking agency.
\newblock \emph{arXiv preprint arXiv:2505.22648}, 2025.

\bibitem[xAI(2025)]{grok3}
xAI.
\newblock Grok 3 beta — the age of reasoning agents, 2025.
\newblock URL \url{https://x.ai/news/grok-3}.

\bibitem[Xbench-Team(2025)]{xbench}
Xbench-Team.
\newblock Xbench-deepsearch, 2025.
\newblock URL \url{https://xbench.org/agi/aisearch}.

\bibitem[Yang et~al.(2025)Yang, Li, Yang, Zhang, Hui, Zheng, Yu, Gao, Huang, Lv, et~al.]{yang2025qwen3}
An~Yang, Anfeng Li, Baosong Yang, Beichen Zhang, Binyuan Hui, Bo~Zheng, Bowen Yu, Chang Gao, Chengen Huang, Chenxu Lv, et~al.
\newblock Qwen3 technical report.
\newblock \emph{arXiv preprint arXiv:2505.09388}, 2025.

\bibitem[Zeng et~al.(2025)Zeng, Lv, Zheng, Hou, Chen, Xie, Wang, Yin, Zeng, Zhang, et~al.]{zeng2025glm}
Aohan Zeng, Xin Lv, Qinkai Zheng, Zhenyu Hou, Bin Chen, Chengxing Xie, Cunxiang Wang, Da~Yin, Hao Zeng, Jiajie Zhang, et~al.
\newblock Glm-4.5: Agentic, reasoning, and coding (arc) foundation models.
\newblock \emph{arXiv preprint arXiv:2508.06471}, 2025.

\bibitem[Zheng et~al.(2025)Zheng, Liu, Li, Chen, Yu, Gao, Dang, Liu, Men, Yang, et~al.]{zheng2025group}
Chujie Zheng, Shixuan Liu, Mingze Li, Xiong-Hui Chen, Bowen Yu, Chang Gao, Kai Dang, Yuqiong Liu, Rui Men, An~Yang, et~al.
\newblock Group sequence policy optimization.
\newblock \emph{arXiv preprint arXiv:2507.18071}, 2025.

\bibitem[Zhou et~al.(2025)Zhou, Leon, Ying, Zhang, Shao, Ye, Chong, Jin, Xie, Cao, et~al.]{bc_zh}
Peilin Zhou, Bruce Leon, Xiang Ying, Can Zhang, Yifan Shao, Qichen Ye, Dading Chong, Zhiling Jin, Chenxuan Xie, Meng Cao, et~al.
\newblock Browsecomp-zh: Benchmarking web browsing ability of large language models in chinese.
\newblock \emph{arXiv preprint arXiv:2504.19314}, 2025.

\end{thebibliography}
\bibliographystyle{colm2024_conference}


\end{document}